\documentclass[11pt,letterpaper]{iffyuan}

\usepackage[all]{hypcap}
\usepackage{algorithm}
\usepackage{algorithmicx}
\usepackage{algpseudocode}
\usepackage{float}
\usepackage{bigstrut}

\usepackage{mathtools}
\usepackage{mathrsfs}
\usepackage{nicefrac}
\usepackage{dsfont}
\usepackage{subcaption}
\expandafter\def\csname ver@subfig.sty\endcsname{}

\usepackage{wrapfig}
\usepackage{lipsum}
\usepackage{stackengine}
\usepackage{adjustbox}
\usepackage{rotating}
\usepackage{makecell}
\usepackage{xspace}
\usepackage{tikz}
\usepackage{array}
\usepackage{ulem}

\newcommand*\circled[1]{\tikz[baseline=(char.base)]{\node[shape=circle,fill=black,text=white,draw,inner sep=.1pt] (char) {#1};}}

\newcommand{\alg}{FSD\xspace}
\newcommand{\scot}{SrCoT\xspace}

\definecolor{blanchedalmond}{rgb}{1.0, 0.92, 0.8}
\definecolor{carmine}{rgb}{0.59, 0.0, 0.09}
\definecolor{lightblue}{rgb}{0.22,0.45,0.70}%

\renewcommand{\mathbf}{\boldsymbol}

\makeatletter
\def\Ddots{\mathinner{\mkern1mu\raise\p@
\vbox{\kern7\p@\hbox{.}}\mkern2mu
\raise4\p@\hbox{.}\mkern2mu\raise7\p@\hbox{.}\mkern1mu}}
\makeatother

\definecolor{amaranth}{rgb}{0.9, 0.17, 0.31}
\definecolor{antiquebrass}{rgb}{0.8, 0.58, 0.46}
\definecolor{antiquefuchsia}{rgb}{0.57, 0.36, 0.51}
\definecolor{chromeyellow}{rgb}{0.31, 0.47, 0.26}

\newtcolorbox{AIbox}[2][]{aibox,title=#2,#1}
\definecolor{lightblue}{rgb}{0.22,0.45,0.70}%
\definecolor{Gray}{gray}{0.95}
\definecolor{Cornsilk}{rgb}{1.0, 0.97, 0.86}

\title{From Seeing to Doing: Bridging Reasoning and Decision for Robotic Manipulation}

\runningtitle{From Seeing to Doing: Bridging Reasoning and Decision for Robotic Manipulation}

\author[1]{Yifu Yuan}
\author[1,\textdagger]{Haiqin Cui}
\author[1,\textdagger]{Yibin Chen}
\author[1]{Zibin Dong}
\author[1]{Fei Ni}
\author[1]{Longxin Kou}
\author[1]{Jinyi Liu}
\author[1]{Pengyi Li}
\author[1]{Yan Zheng}
\author[1]{Jianye Hao}

\affiliation[1]{Tianjin University}

\contribution[\textdagger]{Equal Contribution}

\correspondingauthor{Jianye Hao; Email \href{mailto:jianye.hao@tju.edu.cn}{jianye.hao@tju.edu.cn}}

\setteamlogo{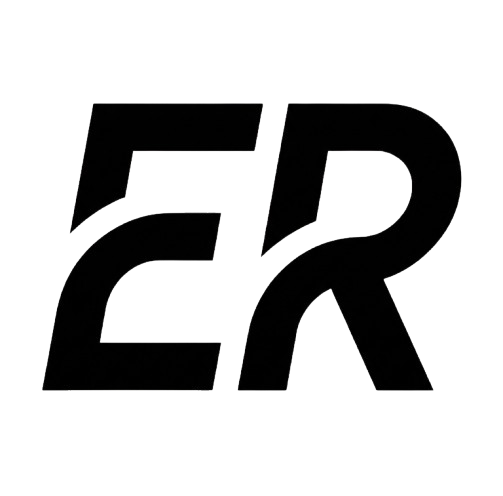}

\abstract{%
Achieving generalization in robotic manipulation remains a critical challenge, particularly for unseen scenarios and novel tasks. Current Vision-Language-Action (VLA) models, while building on top of general Vision-Language Models (VLMs), still fall short of achieving robust zero-shot performance due to the scarcity and heterogeneity prevalent in embodied datasets. To address these limitations, we propose \textbf{\alg} (\textit{From Seeing to Doing}), a novel vision-language model that generates intermediate representations through spatial relationship reasoning, providing fine-grained guidance for robotic manipulation. Our approach combines a hierarchical data construction pipeline for training with a self-consistency mechanism that aligns spatial coordinates with visual signals. Through extensive experiments, we comprehensively validated \alg's capabilities in both ``\textit{seeing}'' and ``\textit{doing}'', achieving outstanding performance across 8 benchmarks for general spatial reasoning and embodied reference abilities, as well as on our proposed more challenging benchmark \textbf{VABench}. We also verified zero-shot capabilities in robot manipulation, demonstrating significant performance improvements over baseline methods in both SimplerEnv and real robot settings. Experimental results show that FSD achieves 40.6\% success rate in SimplerEnv and 72\% success rate across 8 real-world tasks, outperforming the strongest baseline by 30\%. More visualizations and datasets are available on \href{https://embodied-fsd.github.io/}{website}.%
}

\metadata[\faGlobe\ Projects]{\href{https://embodied-fsd.github.io/}{https://embodied-fsd.github.io/}}
\metadata[\raisebox{-1.5pt}{\includegraphics[height=1.03em]{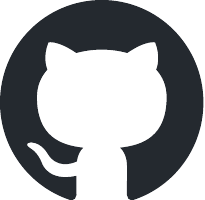}}\ Code]{\href{https://github.com/pickxiguapi/Embodied-FSD}{https://github.com/pickxiguapi/Embodied-FSD}}
\metadata[\raisebox{-1.5pt}{\includegraphics[height=0.96em]{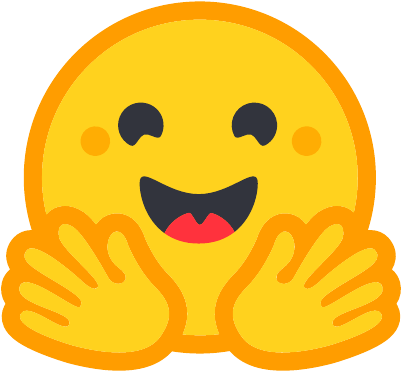}}\ Datasets]{\href{https://huggingface.co/collections/IffYuan/fsd}{https://huggingface.co/collections/IffYuan/fsd}}
\metadata[\faEnvelope\ Contact]{\href{mailto:yuanyf@tju.edu.cn}{yuanyf@tju.edu.cn}}
\metadata[]{\textbf{Published as a conference paper at ICLR 2026}}

\begin{document}

\maketitle
\vspace{3mm}
\vspace{-4mm}
\section{Introduction}

A driving force behind robotics research is the pursuit of generalization: creating agents capable of versatile action across diverse robotic platforms, extending beyond familiar tasks, objects, and environments while adapting to dynamic visual inputs. Current approaches~\cite {kim2024openvla, brohan2023rt} leverage pre-trained Vision-Language Models (VLMs) and transform them into Vision-Language-Action Models (VLAs) using large-scale embodied datasets. This enables systems to interpret natural language instructions and generate robotic manipulation actions~\cite{black2024pi_0}. The intention is to capitalize on the generalization capabilities of VLMs pre-trained on internet-scale data, with the hope that resulting VLAs will adapt to novel scenarios involving unseen objects and tasks. However, empirical evidence~\cite{zheng2024tracevla, zawalski2024robotic, liu2024rdt} demonstrates that directly applying the generalization power of VLMs falls short of achieving strong zero-shot performance on completely novel tasks.

\begin{figure*}[t]
\begin{center}
\centerline{\includegraphics[width=1.0\textwidth]{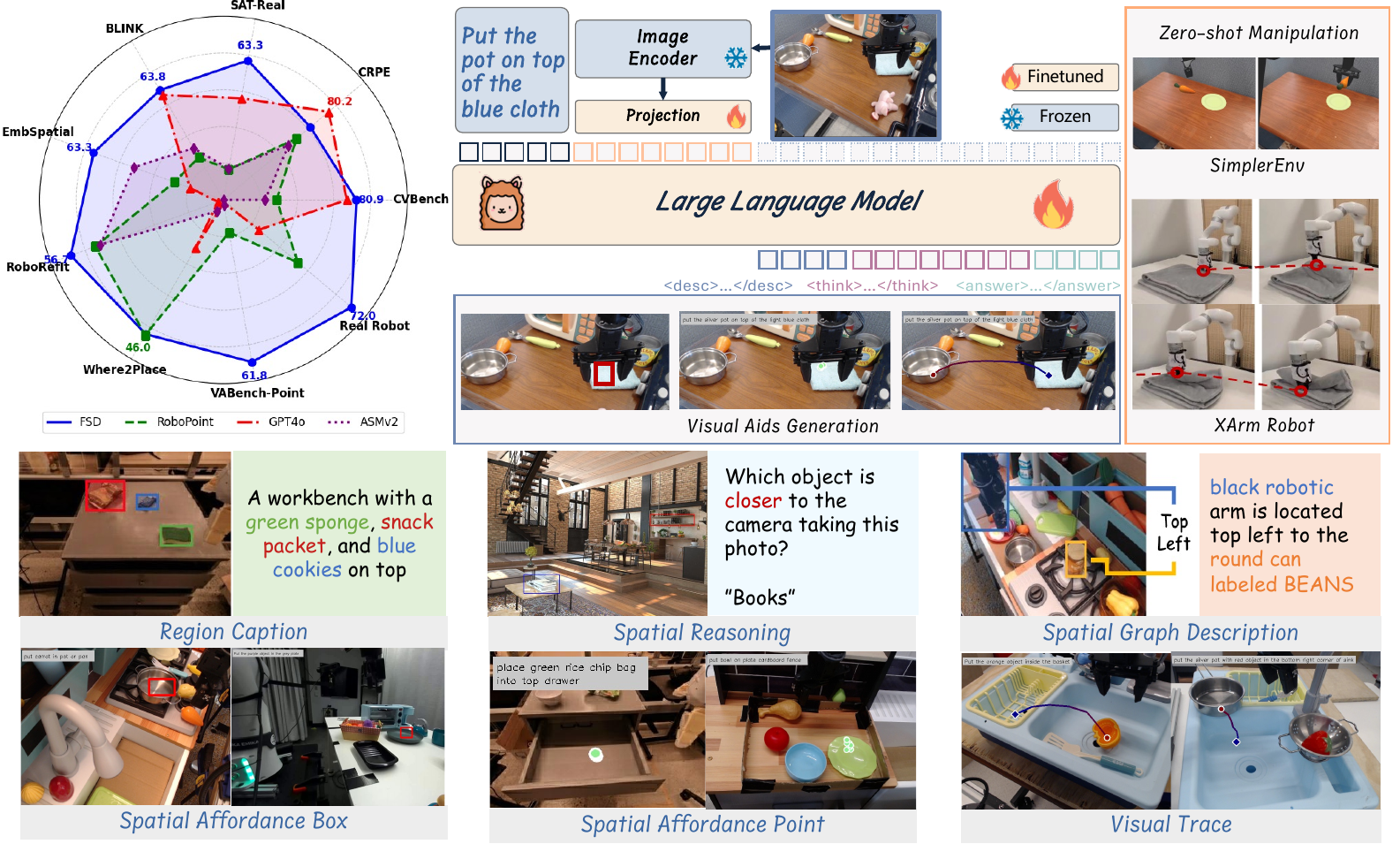}}
\caption{\textbf{Overview of \alg}. FSD unlocks visual aids reasoning and generation through Spatial Relationship Focused CoT, demonstrating exceptional generalization capabilities that enable zero-shot robot manipulation and achieving remarkable performance across multiple benchmarks.}
\label{fig:overview}
\vspace{-28pt}
\end{center}
\end{figure*}

We attribute the limited generalization in VLA-based systems to \textit{scarcity} and \textit{heterogeneity} of datasets. Robotics data remains limited compared to language and vision datasets, preventing similar scaling laws~\cite{kaplan2020scaling, lin2024data}. Despite growth in embodied datasets~\cite{o2023open}, their insufficient coverage and diversity prevent robust zero-shot generalization. 
Additionally, robotic embodiment heterogeneity~\cite{wang2024scaling} causes significant variation in VLA outputs, making end-to-end supervised learning from vision and language to diverse action outputs a potentially unrealistic path toward generalization.

We present a novel pipeline addressing generalization challenges in robotic manipulation. Our approach leverages VLMs' visual understanding capabilities, augmented with step-by-step reasoning to extract unified mid-level structural representations independent of robot embodiment. This representation is key to generalizing learning across diverse physical interactions and dynamic behaviors. Specifically, this mid-level representation includes spatial affordance boxes/points and visual traces, each represented as marked coordinates within visual images. These visual aids provide expressive yet compact spatial information that enables effective reasoning and decisions, overcoming both scarcity and heterogeneity limitations. We introduce \textbf{\alg} (\textit{\textbf{F}rom \textbf{S}eeing to \textbf{D}oing}), a model that generates visual intermediate representations through spatial reasoning (\cref{fig:overview}). \alg comprises three key components: (1) Spatial Relationship-Focused Visual Chain-of-Thought (\textbf{\scot}), which conducts multi-step reasoning anchored by object coordinates and spatial relationships, treating visual aid generation as a reasoning process; (2) A hierarchical data construction pipeline combining large-scale embodied datasets with common sense data, establishing a weak-to-strong capability enhancement training process; and (3) A self-consistency mechanism that aligns understanding and generation by binding spatial coordinates with specific visual signals. To evaluate the accuracy and generalization ability of visual aids generated in complex scenes, we also manually annotated 300 images from both real-world and simulation tasks across various scenarios, forming a \textbf{V}isual \textbf{A}ids Generation Benchmark~(\textbf{VABench}). Through carefully crafted training and evaluation methods, FSD can achieve precise generation of visual aids in different scenarios, then the robot follows spatial affordances and visual traces through simple planning methods to complete action execution.

\alg generalizes effectively to new instructions, and scenes through enhanced reasoning abilities. Our contributions include: (1) A novel paradigm bridging VLM reasoning with embodied decisions via visual aids; (2) The \scot method enabling multi-step reasoning for visual aid generation and guiding zero-shot manipulation; (3) Our weak-to-strong spatial reasoning and visual aids datasets, along with VABench, a manually annotated challenging benchmark for visual aids generation; (4) Superior performance across 8 benchmarks in spatial reasoning, free space reference, and visual aids generation, with zero-shot deployment achieving 40.6\% success in SimplerEnv and 72\% in 8 real-world tasks, outperforming RoboPoint baseline by 30\%.

\label{sec:intro}
\vspace{-1mm}

\section{Related Work}
\label{sec:related_work}
\textbf{Spatial Understanding and Reasoning with VLMs} 
Spatial understanding and reasoning~\cite{liu2023visual, song2024robospatial, du2024embspatial, liao2024reasoning, ray2024sat} require VLMs to infer spatial information beyond 2D RGB images, a capability crucial for embodied AI applications such as navigation~\cite{song2024robospatial, hong20233d, li2024topviewrs} and manipulation~\cite{yuan2024robopoint}. Recent advances include SpatialVLM~\cite{chen2024spatialvlm}, which explicitly incorporates spatial primitives and coordinate systems to enhance geometric reasoning capabilities. Similarly, SpatialRGPT~\cite{cheng2024spatialrgpt} and SpatialBot~\cite{cai2024spatialbot} improve spatial capabilities through more precise spatial relationship modeling. \alg specifically targets embodied manipulation scenarios by enhancing spatial reasoning capabilities through novel \scot mechanisms and sophisticated self-consistency alignment techniques.

\textbf{Visual Chain-of-thought Reasoning} 
The integration of Chain-of-thought (CoT)~\cite{wei2022chain} and its variants~\cite{zhang2022automatic, yao2023beyond, yao2024tree} has significantly enhanced LLM reasoning abilities through structured step-by-step processes. For the multimodal reasoning challenge, researchers have developed various CoT approaches~\cite{mitra2024compositional, zheng2023ddcot, yao2023thinking, wu2025dettoolchain} that establish appropriate reasoning anchors and extend reasoning pathways. Shikra~\cite{chen2023shikra} improves referential expression by incorporating specific visual regions into the reasoning process, while VoCoT~\cite{li2024vocot} extends visually-grounded reasoning chains. VisualCoT~\cite{shao2024visual} provides a comprehensive benchmark validating CoT effectiveness in multi-hop image reasoning tasks. EmbodiedCoT~\cite{zawalski2024robotic} pioneered CoT application in embodied AI by enhancing intermediate reasoning through fine-tuning OpenVLA~\cite{kim2024openvla}. In contrast, \alg uniquely integrates visual-spatial reasoning using spatial relationship graphs as reasoning anchors.

\textbf{Visual Aids Empowered Robotic Manipulation} 
Extracting embodiment-agnostic visual aids to enhance training efficiency has emerged as a promising paradigm in robotic manipulation. Numerous studies~\cite{bharadhwaj2024track2act, wen2023any, xu2024flow, zheng2024tracevla, yuan2024general} have explored robotic policy learning based on visual traces, though most remain confined to specific tasks with cross-embodiment applicability. LLaRVA~\cite{niu2024llarva} advances this field by predicting visual traces to better align visual and action spaces for robot learning, compiling an impressive large visual trace dataset, yet struggles to generalize to novel downstream tasks without task-specific fine-tuning. Spatial affordance represents another effective visual aid, with several works~\cite{yuan2024robopoint, mo2021where2act, qin2020keto, song2024robospatial, ji2025robobrain, yang2025magma, li2025hamster} demonstrating its utility in robotic manipulation tasks. Our key insight is that the scarcity of embodied data fundamentally limits purely data-driven visual aid prediction approaches, impeding zero-shot generalization to unseen scenarios. Therefore, we employ a reasoning-driven paradigm to activate general spatial abilities and enhance generalization.

\section{Bridging Reasoning and Decision through Visual Aids Generation}
\label{method}

To harness VLMs' visual perception capabilities for cross-domain generalization, we introduce \textbf{\alg} (\textit{From Seeing to Doing})—a model exhibiting robust spatial reasoning and generation abilities. We first establish visual aids through spatial affordances and visual traces, then develop Spatial Relationship-Focused CoT (\scot), which leverages object-centric coordinates and their spatial relationships as reasoning anchors. Supporting this approach requires precise spatial understanding and complex instruction-following capabilities to generate coordinate representations. We implement a progressive weak-to-strong data construction pipeline across five capability levels, complemented by a self-consistency alignment mechanism that enhances understanding and generation abilities.

\subsection{Definition of Visual Aids}

\begin{wrapfigure}{r}{0.45\textwidth}
  \vspace{-5px}
  \small
  \centering
  \includegraphics[width=\linewidth]{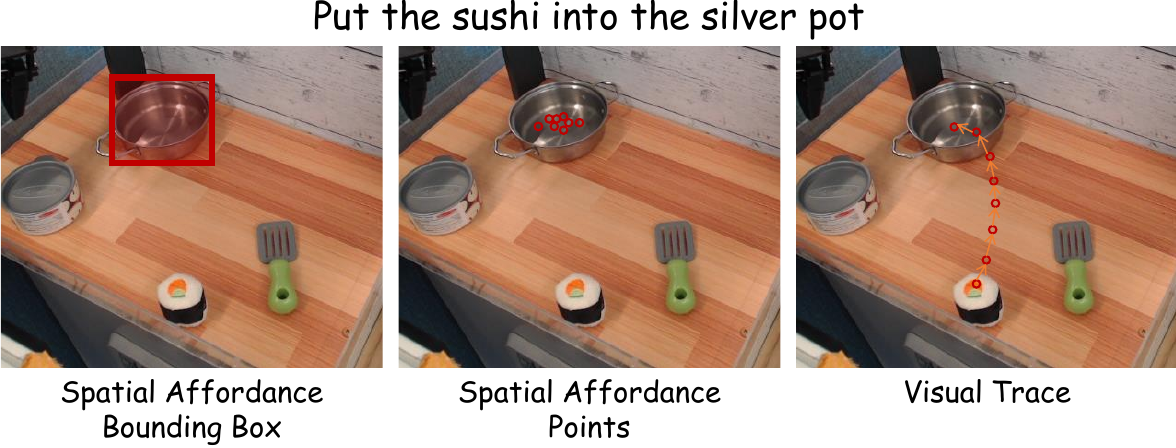}
  \vspace{-10px}
  \caption{Diagrams of Visual Aid Types}
  \label{fig:visual_aids}
\end{wrapfigure}

As shown in \cref{fig:visual_aids}, \alg utilizes three visual aids of increasing complexity, all defined within normalized image coordinates $\mathbf{x} = (p, q) \in [0,1000]^2 \subset \mathbb{R}^2$: \textbf{Spatial affordance boxes} $\mathcal{B} = [x_1, y_1, x_2, y_2]$ define target regions for object placement. For instructions like "\textit{Put the sushi into the silver pot}," the model must infer coordinates for unmarked free space—beyond standard detector capabilities. \textbf{Spatial affordance points} $\mathcal{P} = \{(x_i, y_i) \mid i = 1,2,...,n\}$ provide more precise and flexible placement with reduced redundancy. \textbf{Object-centric visual traces} $\boldsymbol{\tau} = \{\mathbf{x}_t \mid t = 1,2,...,T\}$ represent ordered coordinate sequences that describe manipulation trajectories, where $T$ denotes the sequence length. These traces enable complex instruction execution, cross-embodiment transfer, and collision avoidance. We implement these representations in 2D rather than 3D due to limited high-quality 3D data availability~\cite{zhang2024empowering}. Our object-centric rather than agent-centric visual aids approach effectively circumvents the limitations of heterogeneous embodied data while leveraging general visual datasets without robots, thus enabling robust generalization to novel scenarios and tasks.

\begin{figure}[t]
\begin{center}
\centerline{\includegraphics[width=\textwidth]{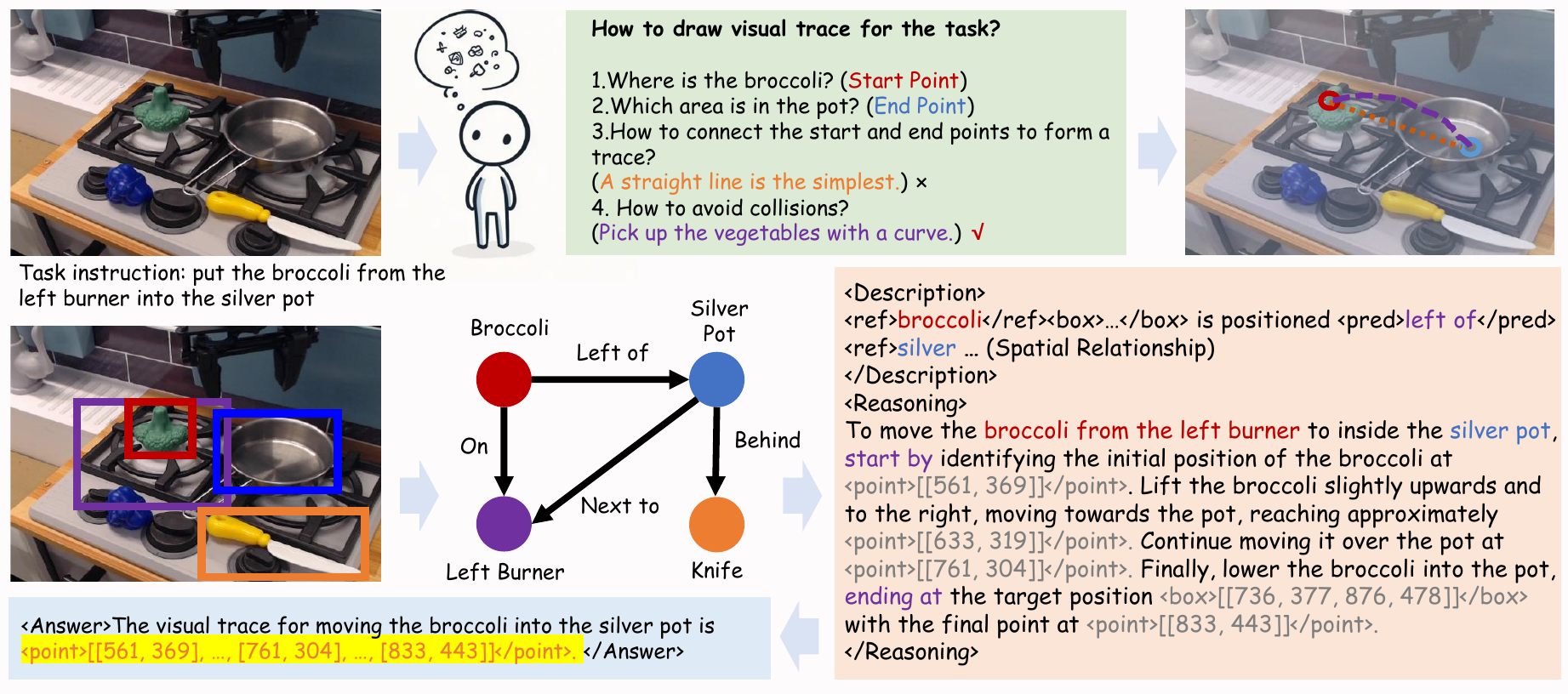}}
\vspace{-5px}
\caption{Inspired by the process of human reasoning, \alg uses a 
spatial relationship graph as an anchor to derive a visual chain-of-thought reasoning process for visual trace generation.}
\label{fig:srcot}
\vspace{-20pt}
\end{center}
\end{figure}

\subsection{Spatial Relationship-Focused Visual Chain-of-thought}

To enable VLMs to generate spatial visual aids, a direct approach is supervised fine-tuning~\cite{niu2024llarva, yuan2024robopoint, li2024llara, zawalski2024robotic} using these aids as a new action space or employing generative models~\cite{shridhar2024generative, xu2024flow}. However, the heterogeneity and scarcity of embodied datasets limit this method. Directly aligning RGB images with coordinate points is challenging and prone to overfitting, hindering generalization. \textit{How can we stimulate VLMs' spatial reasoning abilities to guide the generation rather than merely relying on extensive demonstrations?} Inspired by human cognition (\cref{fig:srcot}~(Top)), when executing tasks like "putting broccoli into a pot," humans first locate relevant objects, then plan movement paths based on relative positions while considering feasibility and obstacles. During this process, humans construct reasoning chains, repeatedly referencing object positions and establishing spatial relationships.

Based on these considerations, we introduce Spatial Relationship-Focused Visual Chain-of-thought (\scot). This approach guides VLMs to generate visual aids through structured reasoning based on spatial relationship graphs. \scot consists of two essential phases: \circled{1} \textbf{Description}: We generate object-centric region captions establishing a task-relevant spatial relationship graph where nodes represent objects with their coordinates and edges denote relative relationships (above, below, left, right, etc.). \circled{2} \textbf{Reasoning}: Using the spatial relationship graph as anchor points, we determine start and end coordinates through object references and free space reasoning, then iteratively derive intermediate points with explicit logical connections between steps. Thus, we prescribe a templated reasoning path for VLMs, enabling \alg to perform analogical reasoning in the spatial domain. While VLMs struggle to directly map future actions to image coordinates, our method leverages known object relationships as reference points for multi-hop analysis, simplifying the reasoning process. \cref{fig:srcot}~(Bottom) demonstrates a complete reasoning sequence. This step-by-step \scot approach, though powerful, fundamentally depends on precise spatial understanding capabilities. To improve the stability and reliability of reasoning paths and reduce model hallucinations, \scot requires the model to generate coordinates in a specified format and bind them to objects while performing object-centric reasoning~\cite{wang2025all, li2024vocot}. We use \textit{<ref>} to mark objects, and \textit{<point>} and \textit{<box>} to mark points and boxes, respectively, ensuring that each object is strictly bound to its corresponding coordinates. This explicit visual-spatial coordinate alignment enhances the \alg’s understanding of the spatial positions and relationships of objects. All coordinates are treated as text and are discretized as integers normalized between 0 and 999.

\subsection{Weak-to-Strong Capability Dataset Construction Pipeline}

The \scot mechanism demands enhanced capabilities from VLMs, including precise reference grounding, spatial understanding, and complex instruction-following capabilities (directly predicting future point trajectories) where mainstream VLMs~\cite{liu2024visual, lin2024vila, ray2024sat, du2024embspatial} show limitations. Consequently, we designed a weak-to-strong data construction pipeline to progressively develop these abilities. 

\textbf{\alg encompasses five hierarchical capability levels:} \circled{1} \textbf{Region Grounding} enables robots to focus on key objects in scenes. Although grounding capabilities have been broadly integrated into current VLMs~\cite{chen2023shikra, you2023ferret}, understanding various small objects and complex scenes for embodied tasks is still limited; \circled{2} \textbf{Spatial Relationship} understanding establishes prerequisite knowledge for spatial reasoning, forming the anchor points for \scot; \circled{3} \textbf{Spatial Reasoning} builds upon these foundations to perform multi-hop analysis of object positions and relationships; and finally, \circled{4} \textbf{Spatial Affordance Generation} and \circled{5} \textbf{Visual Trace Generation} create actionable spatial guidance. Notably, \scot functions as a general visual-spatial reasoning mechanism applicable beyond visual traces to diverse spatial reasoning tasks. Through hierarchical spatial capability training, we enhance VLMs' general spatial reasoning abilities, extending well beyond just embodied domains. 

\begin{figure}[t]
\begin{center}
\centerline{\includegraphics[width=\textwidth]{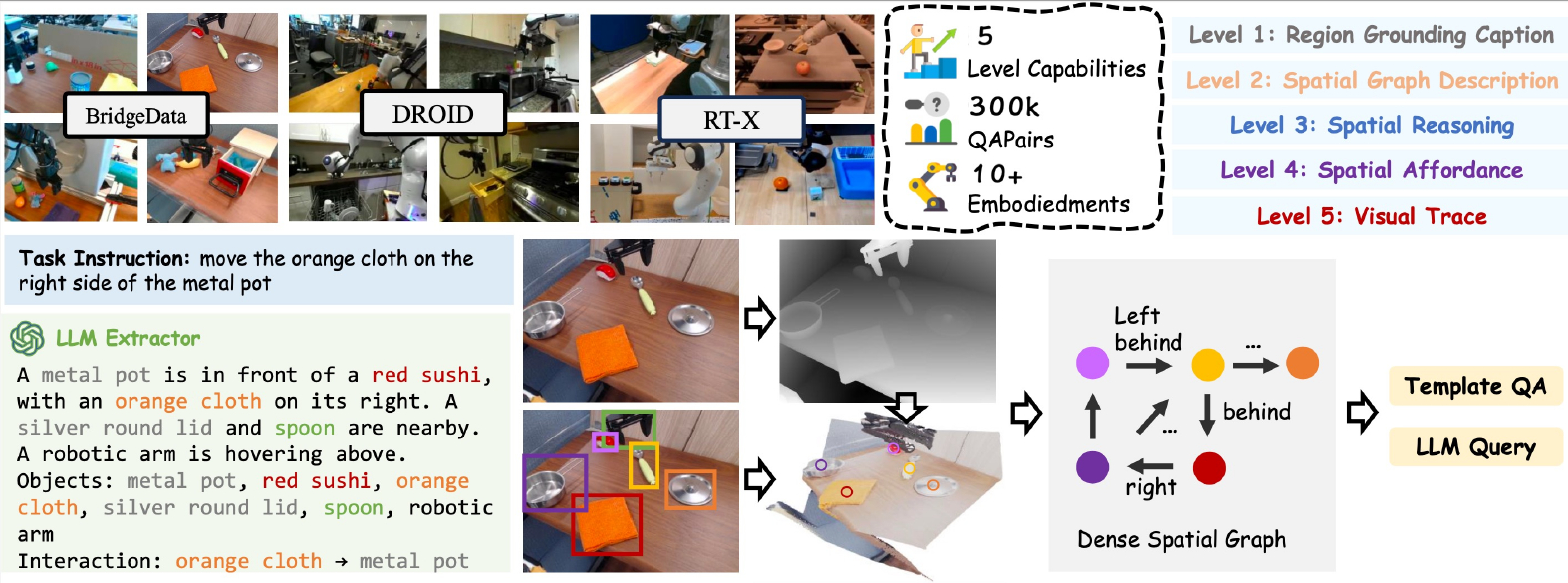}}
\caption{\alg screens data from large-scale embodied datasets, generates ground truth spatial relationship graph. We finally collected 300K data for 10+ embodiments with 5-level capabilities.}
\label{fig:dataset}
\vspace{-18px}
\end{center}
\end{figure}

\textbf{Automatic Dataset Construction:} We leveraged extensive robot data from BridgeDataV2~\cite{walke2023bridgedata}, RT-X~\cite{o2023open}, and Droid~\cite{khazatsky2024droid} to construct FSD's training dataset. Inspired by SpatialVLM~\cite{chen2024spatialvlm} and SpatialRGPT~\cite{cheng2024spatialrgpt}, our automated data construction pipeline created 300K Supervised Fine-Tuning~(SFT) data across numerous formats. After filtering demonstrations with unclear instructions, we used GPT4o~\cite{hurst2024gpt} to nominate task-relevant objects while excluding out-of-range or overly complex items. We then built around objects using GroundedSam~\cite{ren2024grounded} for bounding boxes and segmentation masks (\textbf{Level} \circled{1} dataset). For spatial relationship labels, we reconstructed 3D semantic scene graphs using Metric3Dv2~\cite{hu2024metric3d} for depth estimation, along with WildCamera~\cite{zhu2024tame} and PerspectiveFields~\cite{jin2023perspective} for camera parameters. This enabled 2D-3D mapping and spatial relationship graph construction. (\textbf{Level} \circled{2} dataset). We only generate relative depth sorting data to infer positional relationships, so the accuracy requirement is not high. To improve data quality, in particular, we selected objects with a relative depth gap of 20\% for subsequent generation. Afterward, we randomly sample spatial relationship graphs to construct spatial reasoning QA. We provide image captions, object coordinates, and relationships as a context for GPT4o to create complex QAs (\textbf{Level} \circled{3} Dataset). 

A core aspect of the \alg dataset is the visual aids generation. We employ a simple method with successful human demonstrations from embodied datasets and infer the process from the results. Spatial affordance represents the designated completion area for manipulation tasks. To create spatial affordance labels (\textbf{Level} \circled{4} Dataset), we extract the manipulated object's final position from the terminal frame, combine it with reference object positioning, calculate the precise affordance region, and re-render this information onto the initial frame. For visual trace generation (\textbf{Level} \circled{5} Dataset), we employ a two-stage approach: first applying self-supervised keypoint extraction~\cite{huang2024rekep} to identify grasp points on manipulated objects, then utilizing Cotracker~\cite{karaev2024cotracker3} to capture temporal dynamics from human demonstrations, subsequently projecting these trajectories onto the initial frame. Throughout this process, we employed strict rule-based filters and continually validated our approach against manually annotated test sets, iteratively refining our filtering criteria based on empirical feedback to ensure the resulting dataset met our quality requirements. The dataset presentation, data filtering process, and prompts used to generate the data are provided in \cref{app:weak-to-strong dataset}.

\subsection{Self-Consistent Alignment for Spatial Understanding and Generation} 

High-quality SFT datasets enable VLMs to generate visual aids~\cite{yuan2024robopoint}, yet these models struggle to understand the physical meaning of such annotations since coordinate spaces never appeared in pretraining data. The alignment between image coordinates and actual spatial positions presents a significant challenge. Therefore, we propose a self-consistency mechanism to further align \alg capabilities in spatial understanding and generation. We frame generation tasks inversely as understanding problems: if the forward task requires inferring visual trace $\boldsymbol{\tau}$ from an image $X_v$ and task instruction $X_q$, i.e. $(X_v, X_q) \rightarrow \boldsymbol{\tau}$, we construct the inverse task of predicting possible instructions given an image and visual traces  $(X_v, \boldsymbol{\tau}) \rightarrow X_q$. This bidirectional approach helps the model comprehend spatial coordinates' meanings and aligns coordinate space with image-text modalities, unifying visual aids as both understanding and generation signals while enhancing \alg spatial reasoning capabilities.

\section{Training and Action Execution of \alg}

\textbf{Training:} We follow the instruction tuning pipeline proposed by LLaVA-1.5~\cite{liu2023improved, liu2024visual}. As shown in \cref{fig:overview}, FSD's architecture comprises an image encoder (CLIP-ViT-L-336px~\cite{gao2024clip}), a linear projector, a language tokenizer, and a LLM (Vicuna-13B~\cite{zheng2023judging}). The image encoder processes images into tokens. These visual tokens are then projected into the same embedding space as language tokens through a two-layer linear projector. Only the projector and LLM weights are updated during fine-tuning while the vision encoder and tokenizer remain frozen. We built upon ASMv2~\cite{wang2025all} as our foundation, which already incorporates basic relation conversation and reference grounding capabilities. The training process of FSD is divided into two stages: \textbf{General Spatial Reasoning Enhancement}: Using data from levels 1-3, we focus on improving the model's embodied spatial reasoning capabilities. Following \citet{yuan2024robopoint} and \citet{brohan2023rt}, we discovered that an appropriate data mixture is crucial for downstream performance. Joint training with mixed robotic and internet data ensures the model retains knowledge acquired during pre-training. Consequently, our instruction tuning utilizes a diverse 1.4M sample mixture including general visual question answering (VQA) data. This comprehensive training ensures \alg maintains robust general spatial knowledge while developing embodied capabilities. \textbf{Visual Aids Generation and Understanding}: Using data from levels 4-5 with the self-consistency mechanism, we specifically train visual aids generation and understanding abilities. FSD predicts a fixed set of 8 points for simplification when generating spatial visual traces. Additional training details and the summary of mixture datasets are provided in \cref{app:training details}.

\textbf{Action Execution:} FSD can reason from initial or intermediate task steps, freely selecting needed visual aids. When using bounding boxes, we sample the center as the target point; with affordance points, we directly sample one point. For visual trace execution, we first generate 2D visual traces  $\boldsymbol{\tau}$ and obtain preliminary depth information from depth cameras. Following the pinhole camera model, we employ depth-based back-projection to map these to 3D space, yielding $\boldsymbol{\tau}^{3d} = \{\mathbf{x}_t^{3d} \mid t = 1,2,...,T\}$. Next, based on the spatial position of the first point $\mathbf{x}_1$, we query GraspNet's~\cite{fang2020graspnet} grasp candidates $G$ to match the nearest grasp pose $G^*$. For relatively fixed scenes, we may also use predetermined grasp poses. Subsequently, we optimize the path trajectory using gradient descent-based interpolation, generating complete motion trajectories in SE(3) space, enabling the robotic arm to follow the 3D visual trajectory. When using only spatial affordance, we utilize CuRobo~\cite{sundaralingam2023curobo} as the motion planner to determine execution trajectories $T$ based on the target position $\mathcal{P}$. More details are provided in \cref{app:action execution}. Unlike methods such as LLARVA~\cite{niu2024llarva} and EmbodiedCOT~\cite{zawalski2024robotic} which also utilize visual auxiliary aids, \alg transforms prediction tasks into reasoning tasks, better leveraging visual-spatial common knowledge without requiring scenario-specific fine-tuning.

\section{Visual Aids Generation Benchmark}

Few datasets exist for evaluating visual aid generation, Where2Place~\cite{yuan2024robopoint} provides 100 real-world images from homes and offices but is limited to direct and simple language instructions. And no benchmarks for trajectory prediction. To address this gap, we propose the \textbf{V}isual \textbf{A}ids Generation Benchmark~(\textbf{VABench}). we manually annotated 300 problems from real-world and simulation datasets (OXE~\cite{o2023open}, BridgeData~\cite{walke2023bridgedata}, Droid~\cite{khazatsky2024droid} and Libero~\cite{liu2023libero}), requiring models to infer visual aids given only natural language instructions similar to everyday human commands. For spatial affordance~(\textit{VABench-Point}), we measure the proportion of points falling within target regions. For models that only output bounding boxes, we sample uniformly within boxes. For visual trace~(\textit{VABench-VisualTrace}), we compute MAE and RMSE between ground truth $\boldsymbol{\tau} = \{\mathbf{x}_t \mid t = 1,2,...,T\}$ and predictions $\hat{\boldsymbol{\tau}} = \{\hat{\mathbf{x}}_t \mid t = 1,2,...,\hat{T}\}$: $\text{MAE} = \frac{1}{T} \sum_{t=1}^{T} \|\mathbf{x}_t - \hat{\mathbf{x}}_t\|$, $\text{RMSE} = \sqrt{\frac{1}{T} \sum_{t=1}^{T} \|\mathbf{x}_t - \hat{\mathbf{x}}_t\|^2}$. We interpolate when trajectory lengths differ and normalize coordinates to a 1000×1000 space for consistent evaluation. Since multiple valid solutions exist for each instruction, we simulated human evaluation standards by establishing detailed assessment criteria and added metrics with MLLM-based qualitative scoring (1-10) of visualized trajectories, named GPT Score. We provide detailed dataset information and evaluation procedures in \cref{app:vabench}.

\begin{figure}[t]
\begin{center}
\centerline{\includegraphics[width=0.98\textwidth]{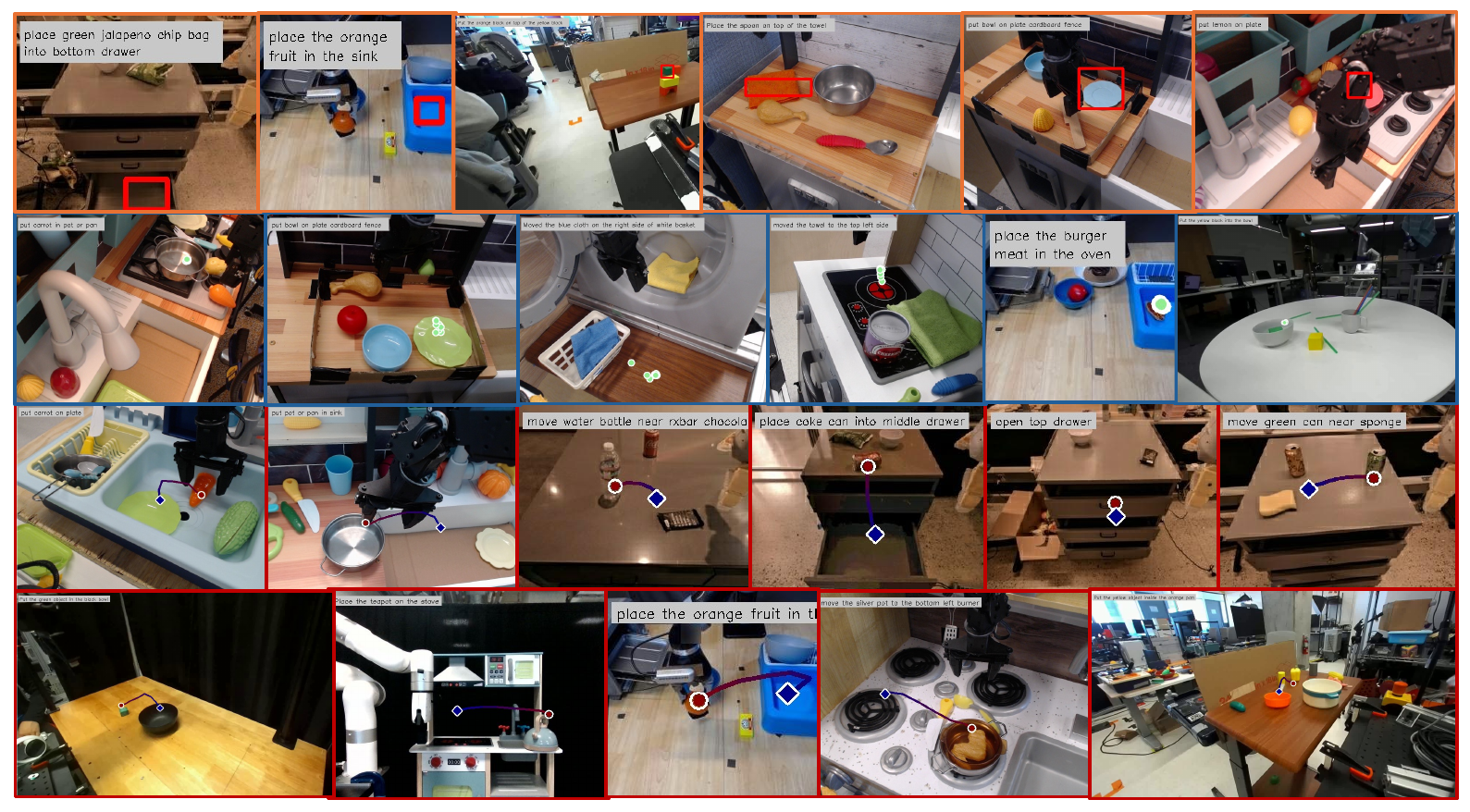}}
\vspace{-5px}
\caption{FSD directly generates visual aids based on task instructions for novel tasks and scenarios. \textit{1st} row: affordance bounding boxes; \textit{2nd} row: affordance points; \textit{3rd} and \textit{4th} rows: visual traces.}
\vspace{-15pt}
\label{fig:visual demo}
\end{center}
\end{figure}

\vspace{-2mm}
\section{Experiments}

We evaluated \alg across two dimensions: \textit{\textbf{Seeing}} and \textit{\textbf{Doing}}. For \textit{Seeing}, we tested its general spatial reasoning and visual aids generation capabilities. For \textit{Doing}, we conducted zero-shot manipulation experiments in both SimplerEnv~\cite{li2024evaluating} simulation and real-world xArm robotic platforms to assess its practical generalization performance.

\begin{table}[t]
\centering
\caption{Performance comparison on 5 spatial reasoning benchmarks. \textbf{Bold} and \underline{underlined} values show best and second-best performance among open-source models.}
\label{tab:general_spatial_benchmark}
\resizebox{\textwidth}{!}{
\begin{tabular}{lcccccccccccccccccc|c}
\toprule
 & \multicolumn{5}{c}{\textbf{CVBench}} & \multicolumn{5}{c}{\textbf{CRPE}} & \multicolumn{2}{c}{\textbf{SAT}} & \multicolumn{5}{c}{\textbf{BLINK}} & \textbf{EmbSp} & \textbf{Rank} \\
\cmidrule(lr){2-6} \cmidrule(lr){7-11} \cmidrule(lr){12-13} \cmidrule(lr){14-18} \cmidrule(lr){19-19} \cmidrule(lr){20-20}
 & Count & 2DRel & 3DDep & 3DDis & Avg. & Exist. & Subj. & Pred. & Obj. & Avg. & Val & Real & Count & MV & RelDepth & SpRel & Avg. & Test \\
\midrule
\rowcolor{gray!10} \multicolumn{20}{l}{\textit{Closed-source models}} \\
GPT-4V & 62.4 & 71.1 & 79.8 & 68.3 & 70.4 & 90.6 & 76.7 & 65.1 & 68.5 & 75.2 & 44.8 & 50.7 & 60.8 & 55.6 & 59.7 & 72.7 & 62.2 & 36.1 & - \\
GPT-4o & 65.9 & 85.5 & 87.8 & 78.2 & 79.4 & 93.3 & 81.9 & 71.8 & 73.6 & 80.2 & 49.4 & 57.5 & 49.2 & 60.2 & 74.2 & 69.2 & 63.2 & 49.1 & - \\
\midrule
\rowcolor{gray!10} \multicolumn{20}{l}{\textit{Open-source models}} \\
LLaVA-1.5-13B & 58.2 & 46.6 & 53.0 & 47.8 & 51.4 & 88.7 & 57.4 & 54.2 & 55.2 & 63.9 & 51.4 & 41.6 & 45.0 & 41.4 & 53.2 & \underline{69.9} & 52.4 & 35.1 & 4.8 \\
SAT-Dynamic-13B & \underline{61.5} & \textbf{89.7} & 80.7 & \underline{73.0} & \underline{76.2} & 87.5 & 60.6 & 57.6 & 65.2 & 67.7 & \textbf{87.7} & \underline{54.9} & 35.8 & \underline{44.4} & \textbf{73.4} & 66.4 & 55.0 & 51.3 & 2.8 \\
RoboPoint-13B & 56.5 & 77.2 & \underline{81.5} & 57.7 & 68.2 & \underline{93.2} & 66.3 & \underline{62.4} & \textbf{70.9} & \underline{73.2} & 53.3 & 46.6 & 48.3 & \underline{44.4} & 62.1 & 65.7 & 55.1 & 51.4 & 2.8 \\
ASMv2-13B & 58.9 & 68.9 & 68.9 & 68.9 & 66.4 & 92.1 & \underline{69.2} & 59.0 & 65.3 & 71.4 & 63.9 & 46.7 & \underline{59.2} & \underline{44.4} & 56.5 & 65.0 & \underline{56.3} & \underline{57.4} & 3.1 \\
\rowcolor{blue!10}
FSD-13B & \textbf{62.4} & \underline{86.5} & \textbf{88.0} & \textbf{86.7} & \textbf{80.9} & \textbf{94.0} & \textbf{75.2} & \textbf{65.1} & \underline{70.4} & \textbf{76.2} & \underline{73.2} & \textbf{63.3} & \textbf{60.0} & \textbf{46.6} & \underline{70.2} & \textbf{78.3} & \textbf{63.8} & \textbf{63.3} & \textbf{1.3} \\
\bottomrule
\end{tabular}
}
\end{table}

\begin{table}[t]
\centering
\caption{Performance comparison on object/free space reference benchmarks. The best results are highlighted.}
\label{tab:object_reference}
\resizebox{\textwidth}{!}{%
\begin{tabular}{l|cccccccc}
\toprule
{\textbf{Benchmark}}
& \textbf{GPT-4o} & \textbf{SpaceLLaVA} & \textbf{LLaVA-NeXT-34B} & \textbf{SpatialBot-3B} & \textbf{ASMv2-13B} & \textbf{RoboBrain-7B} & \textbf{RoboPoint-13B} & \textbf{FSD-13B} \\
\midrule
RoboRefIt & 15.3 & 21.3 & 19.9 & 23.6 & 48.4 & 10.1 & 49.8 & \cellcolor{blue!10} \textbf{56.7} \\
Where2Place & 29.1 & 11.8 & 15.0 & 15.0 & 22.0 & 16.6 & \textbf{46.0} & \cellcolor{blue!10} 45.8 \\
\bottomrule
\end{tabular}%
}
\end{table}

\begin{table}[t]
\small
\centering
\caption{\textbf{Performance comparison on VABench.} The best
results are highlighted in bold.}
\vspace{-5pt}
\label{tab:vabench_combined}
\begin{subtable}[ht]{0.3\textwidth}
\centering
\caption{VABench-Point}
\vspace{-3pt}
\label{tab:vabench_point}
\begin{tabular}{lc}
\toprule
\textbf{Model} & \textbf{Accuracy $\uparrow$} \\
\midrule
GPT4o & 9.30 \\
ASMv2 & 10.07 \\
RoboPoint & 19.09 \\
RoboBrain & 7.00 \\
\rowcolor{blue!10}
FSD & \textbf{61.82} \\
\quad w/o SrCoT & 26.21 \\
\quad w/o Alignment & 55.92 \\
\bottomrule
\end{tabular}
\end{subtable}%
\hfill
\begin{subtable}[ht]{0.6\textwidth}
\centering
\caption{VABench-VisualTrace}
\vspace{-3pt}
\label{tab:vabench_visualtrace}
\begin{tabular}{lccc}
\toprule
\textbf{Model} & \textbf{RMSE$\downarrow$} & \textbf{MAE$\downarrow$} & \textbf{LLM Score$\uparrow$} \\
\midrule
GPT4o & 136.13 & 113.53 & 4.37 \\
DINOv2 Predictor & 128.32 & 117.49 & 4.01 \\
RoboBrain & 121.6 & 103.8 & 4.5 \\
\rowcolor{blue!10}
FSD & \textbf{78.26} & \textbf{63.44} & \textbf{6.21} \\
\quad w/o SrCoT & 99.53 & 80.06 & 5.07 \\
\quad w/o Alignment & 80.48 & 66.80 & 5.92 \\
\bottomrule
\end{tabular}
\end{subtable}
\end{table}

\subsection{Evaluation of Spatial Understanding and Reasoning Capabilities}

\textbf{Benchmarks and Baselines.} \textit{General Spatial Reasoning Capabilities:} We evaluated general spatial reasoning capabilities using five popular benchmarks: CVBench~\cite{tong2024cambrian1}, BLINK~\cite{fu2024blink}, CRPE~\cite{wang2025all}, SAT~\cite{ray2024sat}, and EmbSpatial-Bench~\cite{du2024embspatial}. These benchmarks encompass 15 subtasks measuring various spatial competencies. We included two leading closed-source models: GPT-4o and GPT-4V as performance reference. Subsequently, we conducted comparative analyses against other open-source spatial enhanced MLLMs, including LLaVA-1.5~\cite{liu2023improved}, SAT-Dynamic~\cite{ray2024sat}, RoboPoint~\cite{yuan2024robopoint}, and ASMv2~\cite{wang2025all}, all with 13B parameters. \textit{Object and Free Region Reference Capabilities:} We assessed embodied spatial capabilities using the RoboRefIt~\cite{lu2023vl} and Where2Place~\cite{yuan2024robopoint} benchmarks. We compared mainstream closed-source models and MLLM for enhancing spatial abilities~(SpatialBot~\cite{cai2024spatialbot}, SpaceLLaVA~\cite{chen2024spatialvlm}, and RoboBrain~\cite{ji2025robobrain}). We used the proportion of predicted points within specified regions as the accuracy metric. For models without point output support, we asked models to output bounding boxes of target regions, then sampled evenly within these bounding boxes. \textit{Spatial Visual Aids Capabilities:} We utilized our VABench to evaluate the capabilities. We found few models with this capability for Visual Trace prediction,
so we also trained an end-to-end prediction baseline model using a pre-trained DINOv2~\cite{oquab2023dinov2} encoder coupled with transformer~\cite{vaswani2017attention} architecture to predict visual trajectories, trained on the same visual trajectory data, named \textit{DINOv2 Predictor}. We conducted this comparison to demonstrate the advantages of our reasoning-based FSD approach. More experimental details  are provided in App. \ref{app:benchmarks and baselines}.

\textbf{\alg exhibits superior general spatial reasoning capabilities.}
As shown in \cref{tab:general_spatial_benchmark}, FSD achieves a leading average rank of 1.3 across 15 tasks from spatial benchmarks, significantly outperforming other 13B open-source models and rivaling the closed-source GPT-4o. The model particularly excels in 3D depth perception (88.0\%), distance estimation (86.7\%), and spatial relationship understanding (78.3\%). These results validate our data-centric approach for building robust spatial reasoning, a critical foundation for advanced embodied intelligence.

\textbf{FSD excels in object reference and free space localization.}
The results in \cref{tab:object_reference} demonstrate FSD's ability to accurately identify objects and free spaces from language instructions. For object reference (RoboRefIt), FSD achieves 56.7\% accuracy, surpassing both GPT-4o (15.3\%) and specialist models like RoboPoint (49.8\%) by significant margins. On the more challenging free space reference task (Where2Place), FSD performs competitively with RoboPoint while substantially outperforming other models. This improvement stems from enhanced spatial understanding through our SrCoT mechanism. See App. \ref{app:visualizations} for more visualizations.

\begin{figure}[t]
\begin{center}
\centerline{\includegraphics[width=\textwidth]{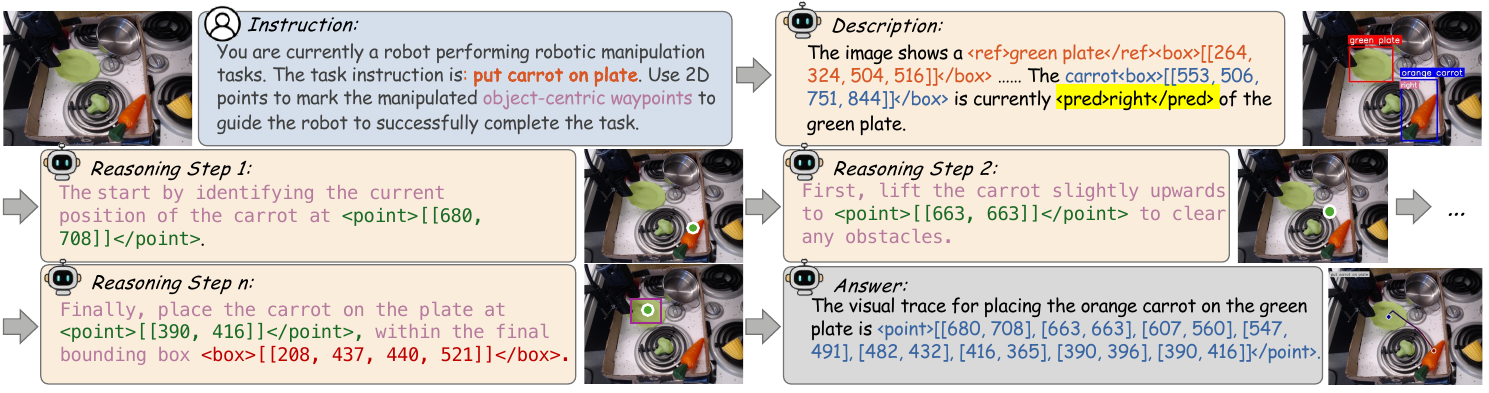}}
\vspace{-5px}
\caption{\textbf{Example of the FSD Reasoning Process} FSD performs point-by-point reasoning and localization, ultimately generating a feasible visual trace.}
\label{fig:one fsd example}
\end{center}
\vspace{-15pt}
\end{figure}

\textbf{FSD demonstrates breakthrough performance in visual aid generation.} As shown in \cref{tab:vabench_combined}, \alg significantly outperforms all baselines in generating precise spatial affordances and visual traces. Specifically, FSD achieves 61.82\% accuracy on VABench-P, over 3x higher than RoboPoint (19.09\%) and attains significantly lower error rates with a better LLM Score on VABench-V. Ablation studies validate the critical contributions of both SrCoT and self-consistency alignment, confirming that our reasoning-based approach enables more accurate predictions than purely data-driven methods. A detailed comparison with RoboBrain~\cite{ji2025robobrain} is available in App. \ref{app:fsd vs robobrain}.

\textbf{Visualization of visual aids generated by FSD}. We visualize the visual aids generated by FSD in \cref{fig:visual demo}, showcasing its effective adaptation to diverse scenes, perspectives, and tasks. \cref{fig:one fsd example} illustrates how the CoT reasoning guides the generation process. Additional visualizations and complete output examples are available in App. \ref{app:visualizations}.

\subsection{Evaluation of the Decision Capability}

\begin{table}[!tbp]
    \centering
    \caption{\textbf{SimplerEnv Evaluation on WidowX Robot.} Each task is tested 24 episodes. Most of the results for end-to-end VLAs are sourced from \citet{chen2025internvlam1spatiallyguidedvisionlanguageaction}, while the results for the remaining models are reproduced in accordance with the official code.}
    \label{tab:simplerenv_widowx}

    \setlength{\tabcolsep}{4pt}

    \resizebox{\textwidth}{!}{%
        \begin{tabular}{llccccc} 
            \toprule
            Type & Model &
            \makecell[b]{Put Spoon \\ on Towel} &
            \makecell[b]{Put Carrot \\ on Plate} &
            \makecell[b]{Stack Green Block \\on Yellow Block} &
            \makecell[b]{Put Eggplant \\ in Yellow Basket} &
            Avg \\
            \midrule

            \multirow{5}{*}{\makecell[l]{End-to-end \\ VLA}}
            & Octo~\cite{team2024octo}         & 41.7 & 8.2 & 0.0 & 56.7 & 26.7 \\
            & $\pi_0$~\cite{black2024pi_0}       & 29.1 & 0.0 & 16.6 & 62.5 & 27.1 \\
            & $\pi_0$-fast~\cite{pertsch2025fast}  & 29.1 & 21.9 & 10.8 & 66.6 & \textbf{48.3} \\
            & OpenVLA~\cite{kim2024openvla}       & 4.2 & 0.0 & 0.0 & 16.7 & 5.2 \\
            & OpenVLA-OFT~\cite{kim2025fine}   & 34.2 & 30.0 & 30.0 & \textbf{72.5} & 41.8 \\
            \midrule

            Modular VLA & MOKA~\cite{liu2024moka} & \textbf{45.8} & 41.6 & \textbf{33.3} & 12.5 & 33.3 \\
            \midrule

            \multirow{2}{*}{\makecell[l]{Affordance \\ VLA}}
            & RoboPoint~\cite{yuan2024robopoint}     & 16.7 & 20.8 & 8.3 & 25.0 & 17.7 \\
            & FSD           & 41.6 & \textbf{50.0} & \textbf{33.3} & 37.5 & 40.6 \\
            \bottomrule
        \end{tabular}%
    }
\end{table}

\begin{figure}[h]
\begin{center}
\centerline{\includegraphics[width=\textwidth]{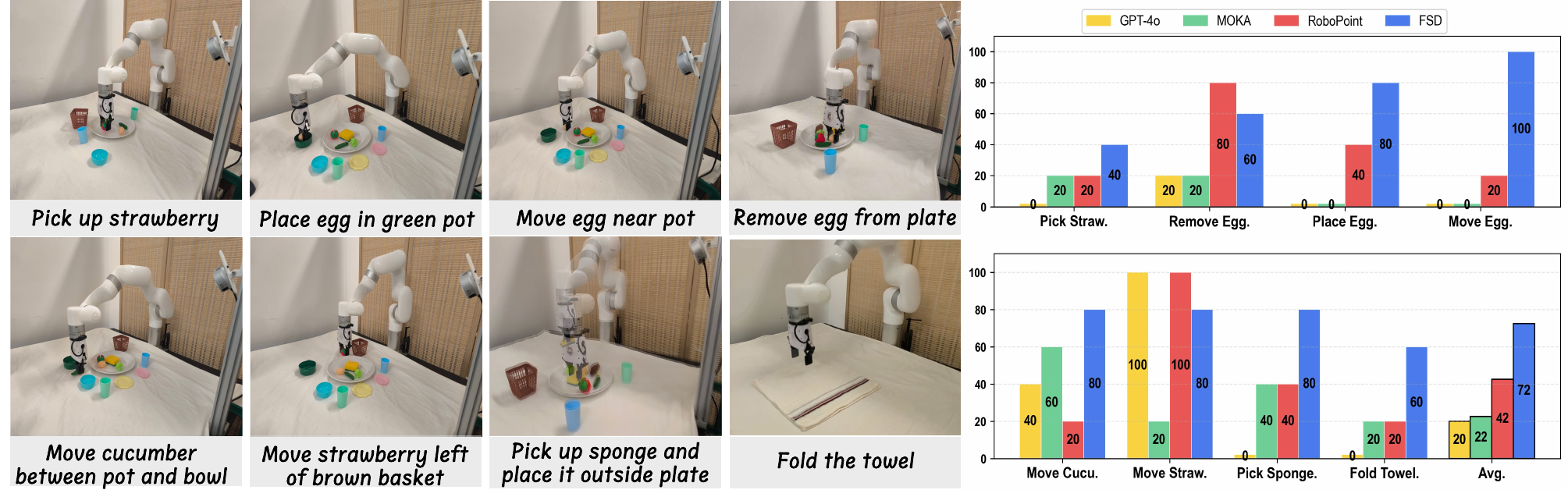}}
\vspace{-5px}
\caption{Real-world robotic manipulation tasks performed by an xArm 6 robot (left) and performance comparison of FSD against baseline models GPT-4o, MOKA, and RoboPoint (right).}
\label{fig:real robot}
\end{center}
\end{figure}

\textbf{SimplerEnv Evaluation.} To assess the zero-shot generalization capability of \alg, we deployed this approach on the WidowX robotic arm and conducted experiments using SimplerEnv~\cite{li2024evaluating}. Since end-to-end VLA methods require fine-tuning, for fair comparison, we ensured that all end-to-end baseline methods were trained using the BridgeData~\cite{walke2023bridgedata} dataset and evaluated under the same settings. We benchmarked mainstream VLAs, including end-to-end models (Octo, the $\pi_0$ series, OpenVLA), the modular-based method MOKA~\cite{liu2024moka}, and the affordance-based RoboPoint~\cite{yuan2024robopoint}.
As shown in \cref{tab:simplerenv_widowx}, although the latest end-to-end VLAs (such as $\pi_0$-fast and OpenVLA-OFT) perform well in this limited generalization scenario, \alg still achieves an average success rate of 40.6\%, significantly outperforming standard zero-shot baselines such as RoboPoint (17.7\%).
Without dedicated fine-tuning, end-to-end VLAs may suffer from severe performance breakdowns (with success rates approaching zero) when faced with substantial variations in backgrounds and instructions. In contrast, \alg demonstrates robust zero-shot generalization. Therefore, we suggest that an important direction for future research is to explore closed-loop policy VLAs explicitly guided by visual trajectories, in order to combine robust planning capabilities with precise execution abilities.

\textbf{Real World Robot Evaluation} As shown in \cref{fig:real robot}, we conducted zero-shot tests with FSD on an xArm 6 robot across 8 tabletop manipulation tasks. The setup included an Intel RealSense L515 LiDAR camera. To test the capabilities of different visual aids, we used visual trace for the sponge and folding tasks, while affordance points were used for other tasks. We compared against GPT-4o, MOKA, and RoboPoint baselines. Robopoint often makes incorrect predictions in tasks involving spatial understanding. The primary cause of MOKA failures stems from the cascading errors of multiple submodules. Under zero-shot conditions, FSD achieved 72\% success rate, outperforming the strongest baseline by more than 30\%. Notably, FSD successfully completed complex tasks with visual trace generation, e.g. towel folding, which was beyond baseline capabilities. Full results are presented in \cref{fig:real robot}. We refer to App. \ref{app:real world experiements} for detailed setup and visualization.

\begin{figure}[h]
    \centering
    \includegraphics[width=\linewidth]{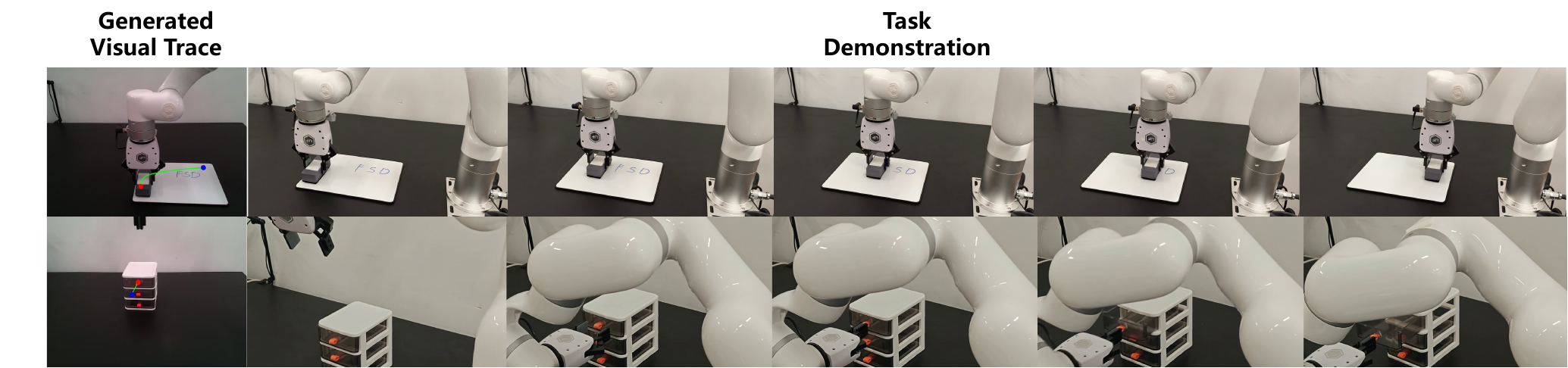}
    \caption{\textbf{Qualitative results on challenging manipulation scenarios}. We evaluate FSD on a contact-rich task (Top: Wiping the whiteboard) and an articulated object task (Bottom: Opening the drawer). The leftmost column visualizes the visual trace generated by FSD, followed by key frames of the real-world execution.}
    \label{fig:challenging}
\end{figure}

\textbf{Qualitative Experiments on Challenging Tasks} We conducted a preliminary exploration and attempted to assess FSD's capability in handling scenarios that require continuous contact and kinematic understanding. We introduced two new challenging tasks: ``Wiping the whiteboard'' and ``opening the drawer.'' As shown in \cref{fig:challenging}, FSD was able to infer the visual traces required for these tasks. For the contact-rich whiteboard task, the model generated a smooth wiping trajectory across the surface. For the articulated drawer task, the model successfully predicted the arc trajectory needed to pull the handle. Although the downstream action controller requires further adjustment and optimization for these more challenging tasks, these qualitative results verify the potential of FSD to bridge perception and complex control.

\begin{wrapfigure}{r}{5.5cm}
    \vspace{-15pt}
    \centering
    \small
    \captionof{table}{Real-world Execution Time}
    \vspace{-8pt}
    \label{tab:execution_time_vertical}
    \begin{tabular}{@{}lcc@{}}
    \hline
    \textbf{Model} & \textbf{Sponge} & \textbf{Cucumber} \\ \hline
    OpenVLA (FT)   & 23s & 12s \\
    MOKA           & 28s & 18s \\
    \textbf{FSD}     & \textbf{14s} & \textbf{10s} \\ \hline
    \end{tabular}
\end{wrapfigure}

\textbf{Real World Execution Latency} We evaluated FSD's execution latency against the end-to-end OpenVLA and modular MOKA pipelines, with results reported in \cref{tab:execution_time_vertical}. FSD achieves the lowest latency due to its single-pass, single-model architecture that generates the entire visual trace in a zero-shot manner. This approach avoids both the costly fine-tuning and step-by-step inference required by OpenVLA and the system overhead of a multi-component pipeline like MOKA. FSD thus effectively balances powerful zero-shot manipulation capabilities with practical, low real-world latency. We primarily use open-loop control, but \alg can also be integrated into a hierarchical learning-based pipeline, using its visual trace to guide and improve closed-loop control. See App. \ref{app:additional experiments} for more details.

\begin{figure}[!t]
\centering
\includegraphics[width=0.97\linewidth]{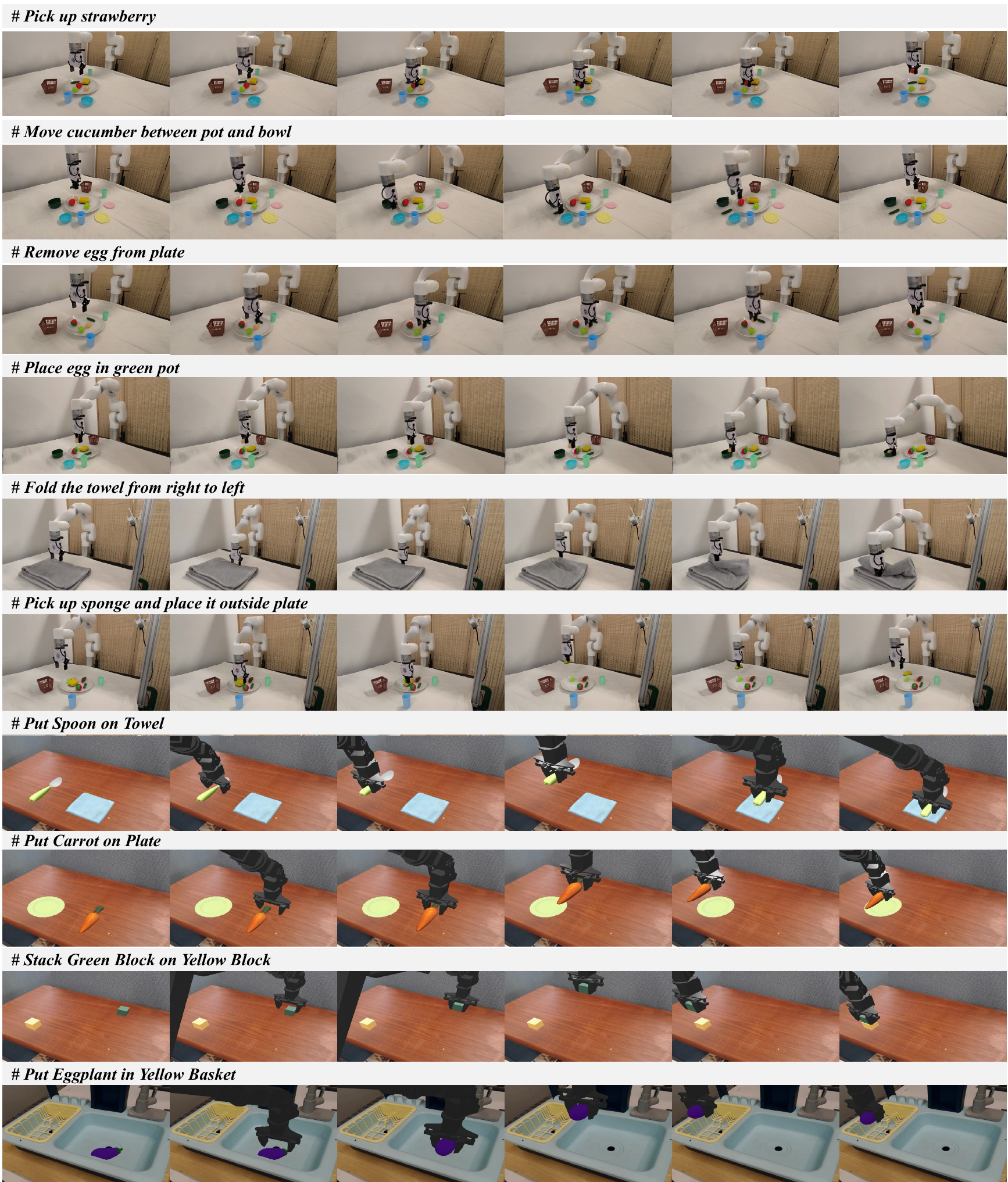}
\caption{\textbf{Visualization of FSD executing tasks}. The first six rows are real-world experiments, and the last four rows are from SIMPLEREnv.}
\label{fig:montage}
\end{figure}

\begin{figure}[t]
    \centering
    \includegraphics[width=0.85\linewidth]{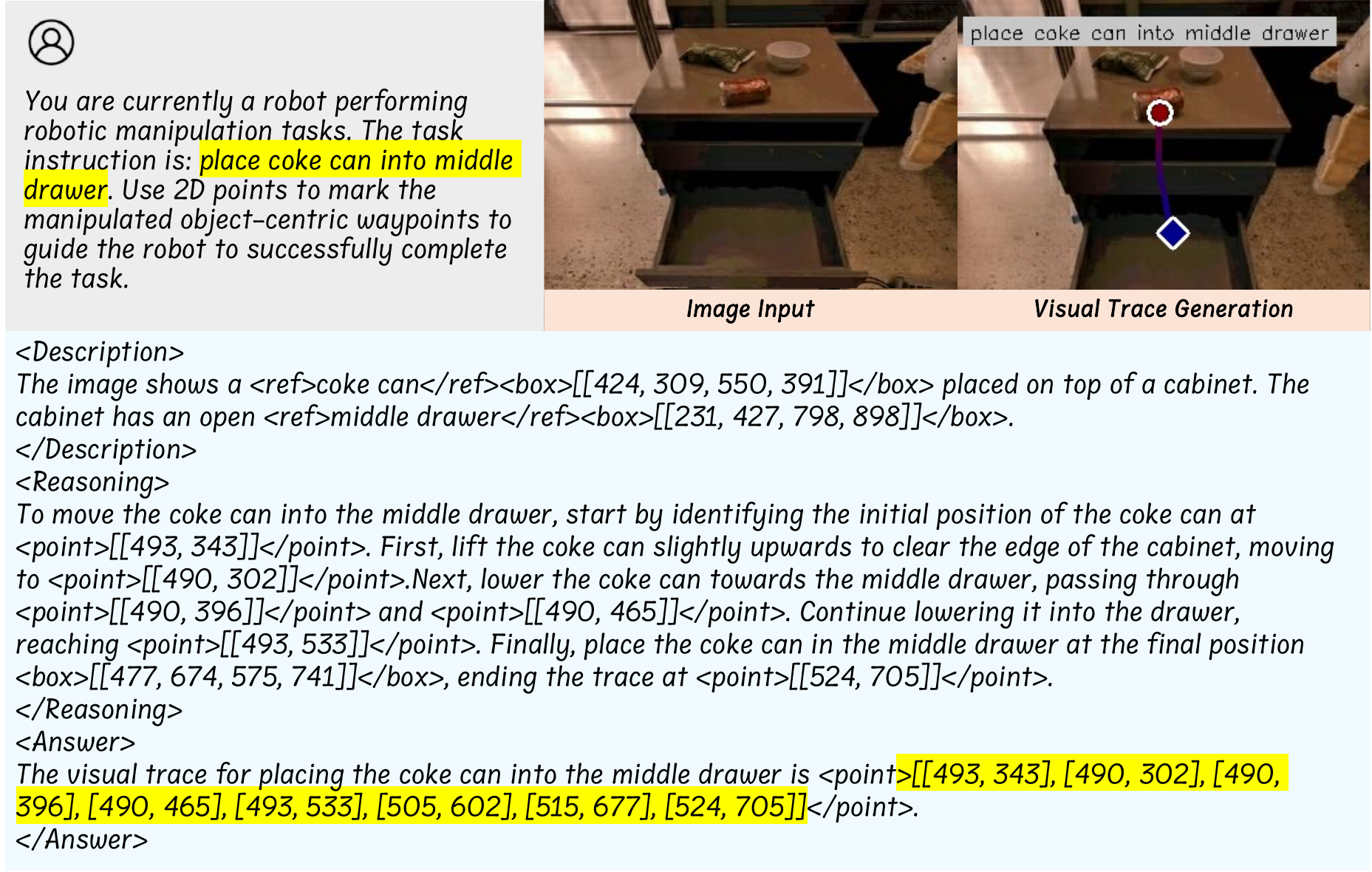}
    \caption{An example and visualization of FSD for generating visual trace.}
    \label{fig:visual_trace_output}
\end{figure}

\begin{figure}[!ht]
    \centering
    \includegraphics[width=0.9\linewidth]{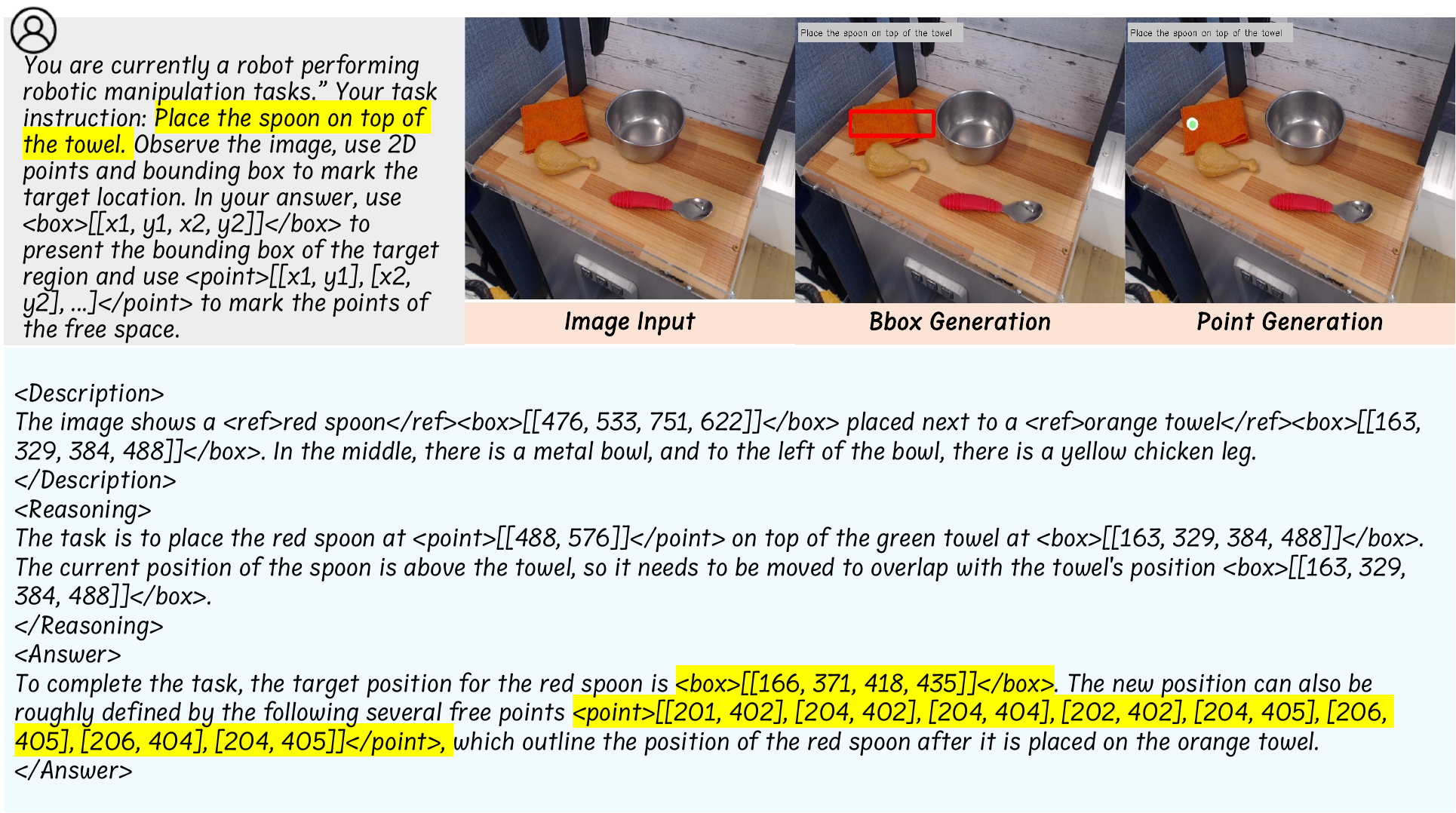}
    \caption{An example and visualization of FSD for generating affordance box and points.}
    \label{fig:point_output}
\end{figure}

\section{Conclusion}
\label{sec:conclusion}

In this paper, we introduced FSD (\textit{From Seeing to Doing}), bridging visual reasoning and robotic manipulation through intermediate spatial representations. Our approach overcomes the critical challenges of scarcity and heterogeneity through three key innovations: a Spatial Relationship-Focused Visual Chain-of-Thought for multi-step reasoning, a hierarchical weak-to-strong data pipeline, and a self-consistency mechanism to align spatial coordinates with visual signals. Experiments demonstrate FSD's superior performance across multiple spatial reasoning and visual aid benchmarks. In zero-shot robotic deployment, FSD achieved an impressive 72\% success rate across diverse tasks, outperforming baselines by 30\%. We acknowledge limitations, such as the reliance on 2D trajectory generation and constraints from training data quality. More limitations and future works are in App.\ref{app:future works}.

\section*{Acknowledgement}

This work is supported by the National Natural Science Foundation of China (Grant Nos. 62422605, 62533021) and the National Natural Science Foundation of China Youth Student Basic Research Project (Grant No. 625B2128). Yifu Yuan is also supported by the Young Science and Technology Scientists Sponsorship Program by CAST - Doctoral Student Special Plan. This work is also supported by the National Natural Science Foundation of China (Grant No. 92370132) and the National Key Research and Development Program of China (Grant No. 2024YFE0210900). Pengyi Li is supported by the Basic Research Project (Grant No. 624B2101).
We would like to thank Zhongwen Xu, Liang Wang, Shuyang Gu, and Han Hu for their participation in the
discussions of this paper and for providing valuable insights. In addition, we would especially like to thank
Yiyang Huang for the constructive suggestions on improving the figures in the manuscript.

\section*{Ethics Statement}

This paper is dedicated to advancing the field of robotic manipulation towards the creation of more effective and versatile robotic assistants. Our research strictly adheres to responsible practices and aligns with the ICLR Code of Ethics. All training data was sourced from large-scale, open-access robotics datasets, with all assets utilized in full compliance with their original licensing and terms of service. We recognize that while the intended applications of this research are positive, the long-term societal impacts of increasing robotic autonomy warrant careful consideration.

\section*{Reproducibility Statement}

To promote transparency and reproducibility within the scientific community, we provide detailed training parameters and resources in the appendix. The complete code for both training and inference has been uploaded to an gitHub repository: \url{https://github.com/pickxiguapi/Embodied-FSD}. We have made datasets and model checkpoints publicly available.

\bibliography{main}

\begin{thebibliography}{80}
\providecommand{\natexlab}[1]{#1}
\providecommand{\url}[1]{\texttt{#1}}
\expandafter\ifx\csname urlstyle\endcsname\relax
  \providecommand{\doi}[1]{doi: #1}\else
  \providecommand{\doi}{doi: \begingroup \urlstyle{rm}\Url}\fi

\bibitem[Bharadhwaj et~al.(2024)Bharadhwaj, Mottaghi, Gupta, and
  Tulsiani]{bharadhwaj2024track2act}
Homanga Bharadhwaj, Roozbeh Mottaghi, Abhinav Gupta, and Shubham Tulsiani.
\newblock Track2act: Predicting point tracks from internet videos enables
  generalizable robot manipulation.
\newblock \emph{arXiv preprint arXiv:2405.01527}, 2024.

\bibitem[Black et~al.(2024)Black, Brown, Driess, Esmail, Equi, Finn, Fusai,
  Groom, Hausman, Ichter, et~al.]{black2024pi_0}
Kevin Black, Noah Brown, Danny Driess, Adnan Esmail, Michael Equi, Chelsea
  Finn, Niccolo Fusai, Lachy Groom, Karol Hausman, Brian Ichter, et~al.
\newblock $\pi_0$: A vision-language-action flow model for general robot
  control.
\newblock \emph{arXiv preprint arXiv:2410.24164}, 2024.

\bibitem[Brohan et~al.(2023)Brohan, Brown, Carbajal, Chebotar, Chen,
  Choromanski, Ding, Driess, Dubey, Finn, et~al.]{brohan2023rt}
Anthony Brohan, Noah Brown, Justice Carbajal, Yevgen Chebotar, Xi~Chen,
  Krzysztof Choromanski, Tianli Ding, Danny Driess, Avinava Dubey, Chelsea
  Finn, et~al.
\newblock Rt-2: Vision-language-action models transfer web knowledge to robotic
  control.
\newblock \emph{arXiv preprint arXiv:2307.15818}, 2023.

\bibitem[Cai et~al.(2024)Cai, Ponomarenko, Yuan, Li, Yang, Dong, and
  Zhao]{cai2024spatialbot}
Wenxiao Cai, Iaroslav Ponomarenko, Jianhao Yuan, Xiaoqi Li, Wankou Yang, Hao
  Dong, and Bo~Zhao.
\newblock Spatialbot: Precise spatial understanding with vision language
  models.
\newblock \emph{arXiv preprint arXiv:2406.13642}, 2024.

\bibitem[Chen et~al.(2024)Chen, Xu, Kirmani, Ichter, Sadigh, Guibas, and
  Xia]{chen2024spatialvlm}
Boyuan Chen, Zhuo Xu, Sean Kirmani, Brain Ichter, Dorsa Sadigh, Leonidas
  Guibas, and Fei Xia.
\newblock Spatialvlm: Endowing vision-language models with spatial reasoning
  capabilities.
\newblock In \emph{Proceedings of the IEEE/CVF Conference on Computer Vision
  and Pattern Recognition}, pages 14455--14465, 2024.

\bibitem[Chen et~al.(2023)Chen, Zhang, Zeng, Zhang, Zhu, and
  Zhao]{chen2023shikra}
Keqin Chen, Zhao Zhang, Weili Zeng, Richong Zhang, Feng Zhu, and Rui Zhao.
\newblock Shikra: Unleashing multimodal llm's referential dialogue magic.
\newblock \emph{arXiv preprint arXiv:2306.15195}, 2023.

\bibitem[Chen et~al.(2025)Chen, Chen, Fu, Gao, Jia, Jin, Li, Mu, Pang, Qiao,
  Tian, Wang, Wang, Wang, Wang, Wang, Wang, Wei, Wu, Yang, Ye, Yu, Zeng, Zhang,
  Zhang, Zhang, Zheng, Zhou, and
  Zhu]{chen2025internvlam1spatiallyguidedvisionlanguageaction}
Xinyi Chen, Yilun Chen, Yanwei Fu, Ning Gao, Jiaya Jia, Weiyang Jin, Hao Li,
  Yao Mu, Jiangmiao Pang, Yu~Qiao, Yang Tian, Bin Wang, Bolun Wang, Fangjing
  Wang, Hanqing Wang, Tai Wang, Ziqin Wang, Xueyuan Wei, Chao Wu, Shuai Yang,
  Jinhui Ye, Junqiu Yu, Jia Zeng, Jingjing Zhang, Jinyu Zhang, Shi Zhang, Feng
  Zheng, Bowen Zhou, and Yangkun Zhu.
\newblock Internvla-m1: A spatially guided vision-language-action framework for
  generalist robot policy, 2025.
\newblock URL \url{https://arxiv.org/abs/2510.13778}.

\bibitem[Cheng et~al.(2024)Cheng, Yin, Fu, Guo, Yang, Kautz, Wang, and
  Liu]{cheng2024spatialrgpt}
An-Chieh Cheng, Hongxu Yin, Yang Fu, Qiushan Guo, Ruihan Yang, Jan Kautz,
  Xiaolong Wang, and Sifei Liu.
\newblock Spatialrgpt: Grounded spatial reasoning in vision language model.
\newblock \emph{arXiv preprint arXiv:2406.01584}, 2024.

\bibitem[Du et~al.(2024)Du, Wu, Li, Huang, and Wei]{du2024embspatial}
Mengfei Du, Binhao Wu, Zejun Li, Xuanjing Huang, and Zhongyu Wei.
\newblock Embspatial-bench: Benchmarking spatial understanding for embodied
  tasks with large vision-language models.
\newblock \emph{arXiv preprint arXiv:2406.05756}, 2024.

\bibitem[Fang et~al.(2020)Fang, Wang, Gou, and Lu]{fang2020graspnet}
Hao-Shu Fang, Chenxi Wang, Minghao Gou, and Cewu Lu.
\newblock Graspnet-1billion: A large-scale benchmark for general object
  grasping.
\newblock In \emph{Proceedings of the IEEE/CVF conference on computer vision
  and pattern recognition}, pages 11444--11453, 2020.

\bibitem[Fu et~al.(2024)Fu, Hu, Li, Feng, Wang, Lin, Roth, Smith, Ma, and
  Krishna]{fu2024blink}
Xingyu Fu, Yushi Hu, Bangzheng Li, Yu~Feng, Haoyu Wang, Xudong Lin, Dan Roth,
  Noah~A Smith, Wei-Chiu Ma, and Ranjay Krishna.
\newblock Blink: Multimodal large language models can see but not perceive.
\newblock In \emph{European Conference on Computer Vision}, pages 148--166.
  Springer, 2024.

\bibitem[Gao et~al.(2024)Gao, Geng, Zhang, Ma, Fang, Zhang, Li, and
  Qiao]{gao2024clip}
Peng Gao, Shijie Geng, Renrui Zhang, Teli Ma, Rongyao Fang, Yongfeng Zhang,
  Hongsheng Li, and Yu~Qiao.
\newblock Clip-adapter: Better vision-language models with feature adapters.
\newblock \emph{International Journal of Computer Vision}, 132\penalty0
  (2):\penalty0 581--595, 2024.

\bibitem[Hartley and Zisserman(2004)]{HartleyZisserman2004}
Richard Hartley and Andrew Zisserman.
\newblock \emph{Multiple View Geometry in Computer Vision}.
\newblock Cambridge University Press, 2nd edition, 2004.
\newblock ISBN 0521540518.

\bibitem[Hong et~al.(2023)Hong, Zhen, Chen, Zheng, Du, Chen, and
  Gan]{hong20233d}
Yining Hong, Haoyu Zhen, Peihao Chen, Shuhong Zheng, Yilun Du, Zhenfang Chen,
  and Chuang Gan.
\newblock 3d-llm: Injecting the 3d world into large language models.
\newblock \emph{Advances in Neural Information Processing Systems},
  36:\penalty0 20482--20494, 2023.

\bibitem[Hu et~al.(2024)Hu, Yin, Zhang, Cai, Long, Chen, Wang, Yu, Shen, and
  Shen]{hu2024metric3d}
Mu~Hu, Wei Yin, Chi Zhang, Zhipeng Cai, Xiaoxiao Long, Hao Chen, Kaixuan Wang,
  Gang Yu, Chunhua Shen, and Shaojie Shen.
\newblock Metric3d v2: A versatile monocular geometric foundation model for
  zero-shot metric depth and surface normal estimation.
\newblock \emph{arXiv preprint arXiv:2404.15506}, 2024.

\bibitem[Huang et~al.(2024)Huang, Wang, Li, Zhang, and Fei-Fei]{huang2024rekep}
Wenlong Huang, Chen Wang, Yunzhu Li, Ruohan Zhang, and Li~Fei-Fei.
\newblock Rekep: Spatio-temporal reasoning of relational keypoint constraints
  for robotic manipulation.
\newblock \emph{arXiv preprint arXiv:2409.01652}, 2024.

\bibitem[Hurst et~al.(2024)Hurst, Lerer, Goucher, Perelman, Ramesh, Clark,
  Ostrow, Welihinda, Hayes, Radford, et~al.]{hurst2024gpt}
Aaron Hurst, Adam Lerer, Adam~P Goucher, Adam Perelman, Aditya Ramesh, Aidan
  Clark, AJ~Ostrow, Akila Welihinda, Alan Hayes, Alec Radford, et~al.
\newblock Gpt-4o system card.
\newblock \emph{arXiv preprint arXiv:2410.21276}, 2024.

\bibitem[Ji et~al.(2025)Ji, Tan, Shi, Hao, Zhang, Zhang, Wang, Zhao, Mu, An,
  Xue, Su, Lyu, Zheng, Liu, Wang, and Zhang]{ji2025robobrain}
Yuheng Ji, Huajie Tan, Jiayu Shi, Xiaoshuai Hao, Yuan Zhang, Hengyuan Zhang,
  Pengwei Wang, Mengdi Zhao, Yao Mu, Pengju An, Xinda Xue, Qinghang Su, Huaihai
  Lyu, Xiaolong Zheng, Jiaming Liu, Zhongyuan Wang, and Shanghang Zhang.
\newblock Robobrain: A unified brain model for robotic manipulation from
  abstract to concrete.
\newblock \emph{arXiv preprint arXiv:2502.21257}, 2025.

\bibitem[Jin et~al.(2023)Jin, Zhang, Hold-Geoffroy, Wang, Blackburn-Matzen,
  Sticha, and Fouhey]{jin2023perspective}
Linyi Jin, Jianming Zhang, Yannick Hold-Geoffroy, Oliver Wang, Kevin
  Blackburn-Matzen, Matthew Sticha, and David~F Fouhey.
\newblock Perspective fields for single image camera calibration.
\newblock In \emph{Proceedings of the IEEE/CVF Conference on Computer Vision
  and Pattern Recognition}, pages 17307--17316, 2023.

\bibitem[Kaplan et~al.(2020)Kaplan, McCandlish, Henighan, Brown, Chess, Child,
  Gray, Radford, Wu, and Amodei]{kaplan2020scaling}
Jared Kaplan, Sam McCandlish, Tom Henighan, Tom~B Brown, Benjamin Chess, Rewon
  Child, Scott Gray, Alec Radford, Jeffrey Wu, and Dario Amodei.
\newblock Scaling laws for neural language models.
\newblock \emph{arXiv preprint arXiv:2001.08361}, 2020.

\bibitem[Karaev et~al.(2024)Karaev, Makarov, Wang, Neverova, Vedaldi, and
  Rupprecht]{karaev2024cotracker3}
Nikita Karaev, Iurii Makarov, Jianyuan Wang, Natalia Neverova, Andrea Vedaldi,
  and Christian Rupprecht.
\newblock Cotracker3: Simpler and better point tracking by pseudo-labelling
  real videos.
\newblock \emph{arXiv preprint arXiv:2410.11831}, 2024.

\bibitem[Khazatsky et~al.(2024)Khazatsky, Pertsch, Nair, Balakrishna, Dasari,
  Karamcheti, Nasiriany, Srirama, Chen, Ellis, et~al.]{khazatsky2024droid}
Alexander Khazatsky, Karl Pertsch, Suraj Nair, Ashwin Balakrishna, Sudeep
  Dasari, Siddharth Karamcheti, Soroush Nasiriany, Mohan~Kumar Srirama,
  Lawrence~Yunliang Chen, Kirsty Ellis, et~al.
\newblock Droid: A large-scale in-the-wild robot manipulation dataset.
\newblock \emph{arXiv preprint arXiv:2403.12945}, 2024.

\bibitem[Kim et~al.(2024)Kim, Pertsch, Karamcheti, Xiao, Balakrishna, Nair,
  Rafailov, Foster, Lam, Sanketi, et~al.]{kim2024openvla}
Moo~Jin Kim, Karl Pertsch, Siddharth Karamcheti, Ted Xiao, Ashwin Balakrishna,
  Suraj Nair, Rafael Rafailov, Ethan Foster, Grace Lam, Pannag Sanketi, et~al.
\newblock Openvla: An open-source vision-language-action model.
\newblock \emph{arXiv preprint arXiv:2406.09246}, 2024.

\bibitem[Kim et~al.(2025)Kim, Finn, and Liang]{kim2025fine}
Moo~Jin Kim, Chelsea Finn, and Percy Liang.
\newblock Fine-tuning vision-language-action models: Optimizing speed and
  success.
\newblock \emph{arXiv preprint arXiv:2502.19645}, 2025.

\bibitem[Li et~al.(2024{\natexlab{a}})Li, Zhang, Guo, Zhang, Li, Zhang, Zhang,
  Zhang, Li, Liu, et~al.]{li2024llava}
Bo~Li, Yuanhan Zhang, Dong Guo, Renrui Zhang, Feng Li, Hao Zhang, Kaichen
  Zhang, Peiyuan Zhang, Yanwei Li, Ziwei Liu, et~al.
\newblock Llava-onevision: Easy visual task transfer.
\newblock \emph{arXiv preprint arXiv:2408.03326}, 2024{\natexlab{a}}.

\bibitem[Li et~al.(2024{\natexlab{b}})Li, Zhang, Zhou, Collier, Korhonen, and
  Vuli{\'c}]{li2024topviewrs}
Chengzu Li, Caiqi Zhang, Han Zhou, Nigel Collier, Anna Korhonen, and Ivan
  Vuli{\'c}.
\newblock Topviewrs: Vision-language models as top-view spatial reasoners.
\newblock \emph{arXiv preprint arXiv:2406.02537}, 2024{\natexlab{b}}.

\bibitem[Li et~al.(2024{\natexlab{c}})Li, Mata, Park, Kahatapitiya, Jang,
  Shang, Ranasinghe, Burgert, Cai, Lee, et~al.]{li2024llara}
Xiang Li, Cristina Mata, Jongwoo Park, Kumara Kahatapitiya, Yoo~Sung Jang,
  Jinghuan Shang, Kanchana Ranasinghe, Ryan Burgert, Mu~Cai, Yong~Jae Lee,
  et~al.
\newblock Llara: Supercharging robot learning data for vision-language policy.
\newblock \emph{arXiv preprint arXiv:2406.20095}, 2024{\natexlab{c}}.

\bibitem[Li et~al.(2024{\natexlab{d}})Li, Hsu, Gu, Pertsch, Mees, Walke, Fu,
  Lunawat, Sieh, Kirmani, et~al.]{li2024evaluating}
Xuanlin Li, Kyle Hsu, Jiayuan Gu, Karl Pertsch, Oier Mees, Homer~Rich Walke,
  Chuyuan Fu, Ishikaa Lunawat, Isabel Sieh, Sean Kirmani, et~al.
\newblock Evaluating real-world robot manipulation policies in simulation.
\newblock \emph{arXiv preprint arXiv:2405.05941}, 2024{\natexlab{d}}.

\bibitem[Li et~al.(2025)Li, Deng, Zhang, Jang, Memmel, Yu, Garrett, Ramos, Fox,
  Li, et~al.]{li2025hamster}
Yi~Li, Yuquan Deng, Jesse Zhang, Joel Jang, Marius Memmel, Raymond Yu,
  Caelan~Reed Garrett, Fabio Ramos, Dieter Fox, Anqi Li, et~al.
\newblock Hamster: Hierarchical action models for open-world robot
  manipulation.
\newblock \emph{arXiv preprint arXiv:2502.05485}, 2025.

\bibitem[Li et~al.(2024{\natexlab{e}})Li, Luo, Zhang, Qiu, and
  Wei]{li2024vocot}
Zejun Li, Ruipu Luo, Jiwen Zhang, Minghui Qiu, and Zhongyu Wei.
\newblock Vocot: Unleashing visually grounded multi-step reasoning in large
  multi-modal models.
\newblock \emph{arXiv preprint arXiv:2405.16919}, 2024{\natexlab{e}}.

\bibitem[Liao et~al.(2024)Liao, Mahmood, Fidler, and Acuna]{liao2024reasoning}
Yuan-Hong Liao, Rafid Mahmood, Sanja Fidler, and David Acuna.
\newblock Reasoning paths with reference objects elicit quantitative spatial
  reasoning in large vision-language models.
\newblock \emph{arXiv preprint arXiv:2409.09788}, 2024.

\bibitem[Lin et~al.(2024{\natexlab{a}})Lin, Hu, Sheng, Wen, You, and
  Gao]{lin2024data}
Fanqi Lin, Yingdong Hu, Pingyue Sheng, Chuan Wen, Jiacheng You, and Yang Gao.
\newblock Data scaling laws in imitation learning for robotic manipulation.
\newblock \emph{arXiv preprint arXiv:2410.18647}, 2024{\natexlab{a}}.

\bibitem[Lin et~al.(2024{\natexlab{b}})Lin, Yin, Ping, Molchanov, Shoeybi, and
  Han]{lin2024vila}
Ji~Lin, Hongxu Yin, Wei Ping, Pavlo Molchanov, Mohammad Shoeybi, and Song Han.
\newblock Vila: On pre-training for visual language models.
\newblock In \emph{Proceedings of the IEEE/CVF Conference on Computer Vision
  and Pattern Recognition}, pages 26689--26699, 2024{\natexlab{b}}.

\bibitem[Liu et~al.(2023{\natexlab{a}})Liu, Zhu, Gao, Feng, Liu, Zhu, and
  Stone]{liu2023libero}
Bo~Liu, Yifeng Zhu, Chongkai Gao, Yihao Feng, Qiang Liu, Yuke Zhu, and Peter
  Stone.
\newblock Libero: Benchmarking knowledge transfer for lifelong robot learning.
\newblock \emph{Advances in Neural Information Processing Systems},
  36:\penalty0 44776--44791, 2023{\natexlab{a}}.

\bibitem[Liu et~al.(2024{\natexlab{a}})Liu, Fang, Abbeel, and
  Levine]{liu2024moka}
Fangchen Liu, Kuan Fang, Pieter Abbeel, and Sergey Levine.
\newblock Moka: Open-vocabulary robotic manipulation through mark-based visual
  prompting.
\newblock In \emph{First Workshop on Vision-Language Models for Navigation and
  Manipulation at ICRA 2024}, 2024{\natexlab{a}}.

\bibitem[Liu et~al.(2023{\natexlab{b}})Liu, Emerson, and
  Collier]{liu2023visual}
Fangyu Liu, Guy Emerson, and Nigel Collier.
\newblock Visual spatial reasoning.
\newblock \emph{Transactions of the Association for Computational Linguistics},
  11:\penalty0 635--651, 2023{\natexlab{b}}.

\bibitem[Liu et~al.(2023{\natexlab{c}})Liu, Li, Wu, and Lee]{liu2023improved}
Haotian Liu, Chunyuan Li, Qingyang Wu, and Yong~Jae Lee.
\newblock Improved baselines with visual instruction tuning.
\newblock \emph{arXiv preprint arXiv:2310.03744}, 2023{\natexlab{c}}.

\bibitem[Liu et~al.(2024{\natexlab{b}})Liu, Li, Li, Li, Zhang, Shen, and
  Lee]{liu2024llavanext}
Haotian Liu, Chunyuan Li, Yuheng Li, Bo~Li, Yuanhan Zhang, Sheng Shen, and
  Yong~Jae Lee.
\newblock Llava-next: Improved reasoning, ocr, and world knowledge,
  2024{\natexlab{b}}.
\newblock URL \url{https://llava-vl.github.io/blog/2024-01-30-llava-next/}.

\bibitem[Liu et~al.(2024{\natexlab{c}})Liu, Li, Wu, and Lee]{liu2024visual}
Haotian Liu, Chunyuan Li, Qingyang Wu, and Yong~Jae Lee.
\newblock Visual instruction tuning.
\newblock \emph{Advances in neural information processing systems}, 36,
  2024{\natexlab{c}}.

\bibitem[Liu et~al.(2023{\natexlab{d}})Liu, Zeng, Ren, Li, Zhang, Yang, Li,
  Yang, Su, Zhu, et~al.]{liu2023grounding}
Shilong Liu, Zhaoyang Zeng, Tianhe Ren, Feng Li, Hao Zhang, Jie Yang, Chunyuan
  Li, Jianwei Yang, Hang Su, Jun Zhu, et~al.
\newblock Grounding dino: Marrying dino with grounded pre-training for open-set
  object detection.
\newblock \emph{arXiv preprint arXiv:2303.05499}, 2023{\natexlab{d}}.

\bibitem[Liu et~al.(2024{\natexlab{d}})Liu, Wu, Li, Tan, Chen, Wang, Xu, Su,
  and Zhu]{liu2024rdt}
Songming Liu, Lingxuan Wu, Bangguo Li, Hengkai Tan, Huayu Chen, Zhengyi Wang,
  Ke~Xu, Hang Su, and Jun Zhu.
\newblock Rdt-1b: a diffusion foundation model for bimanual manipulation.
\newblock \emph{arXiv preprint arXiv:2410.07864}, 2024{\natexlab{d}}.

\bibitem[Lu et~al.(2023)Lu, Fan, Deng, Liu, Li, and Wang]{lu2023vl}
Yuhao Lu, Yixuan Fan, Beixing Deng, Fangfu Liu, Yali Li, and Shengjin Wang.
\newblock Vl-grasp: a 6-dof interactive grasp policy for language-oriented
  objects in cluttered indoor scenes.
\newblock In \emph{2023 IEEE/RSJ International Conference on Intelligent Robots
  and Systems (IROS)}, pages 976--983. IEEE, 2023.

\bibitem[Mitra et~al.(2024)Mitra, Huang, Darrell, and
  Herzig]{mitra2024compositional}
Chancharik Mitra, Brandon Huang, Trevor Darrell, and Roei Herzig.
\newblock Compositional chain-of-thought prompting for large multimodal models.
\newblock In \emph{Proceedings of the IEEE/CVF Conference on Computer Vision
  and Pattern Recognition}, pages 14420--14431, 2024.

\bibitem[Mo et~al.(2021)Mo, Guibas, Mukadam, Gupta, and
  Tulsiani]{mo2021where2act}
Kaichun Mo, Leonidas~J Guibas, Mustafa Mukadam, Abhinav Gupta, and Shubham
  Tulsiani.
\newblock Where2act: From pixels to actions for articulated 3d objects.
\newblock In \emph{Proceedings of the IEEE/CVF International Conference on
  Computer Vision}, pages 6813--6823, 2021.

\bibitem[Niu et~al.(2024)Niu, Sharma, Biamby, Quenum, Bai, Shi, Darrell, and
  Herzig]{niu2024llarva}
Dantong Niu, Yuvan Sharma, Giscard Biamby, Jerome Quenum, Yutong Bai, Baifeng
  Shi, Trevor Darrell, and Roei Herzig.
\newblock Llarva: Vision-action instruction tuning enhances robot learning.
\newblock \emph{arXiv preprint arXiv:2406.11815}, 2024.

\bibitem[O'Neill et~al.(2023)O'Neill, Arthurs, Avila Belbute-Peres, Balaguer,
  Bechtle, Bidoia, Burden, Chang, Chen, Davchev, et~al.]{o2023open}
Jake O'Neill, Abraham Arthurs, Fábio Avila Belbute-Peres, Julian Balaguer,
  Sarah Bechtle, Gemma Bidoia, Kyle Burden, Erwin Chang, Sheila Chen, Todor
  Davchev, et~al.
\newblock Open x-embodiment: Robotic learning datasets and rt-x models.
\newblock \emph{arXiv preprint arXiv:2310.08864}, 2023.

\bibitem[Oquab et~al.(2023)Oquab, Darcet, Moutakanni, Vo, Szafraniec, Khalidov,
  Fernandez, Haziza, Massa, El-Nouby, et~al.]{oquab2023dinov2}
Maxime Oquab, Timoth{\'e}e Darcet, Th{\'e}o Moutakanni, Huy Vo, Marc
  Szafraniec, Vasil Khalidov, Pierre Fernandez, Daniel Haziza, Francisco Massa,
  Alaaeldin El-Nouby, et~al.
\newblock Dinov2: Learning robust visual features without supervision.
\newblock \emph{arXiv preprint arXiv:2304.07193}, 2023.

\bibitem[Pertsch et~al.(2025)Pertsch, Stachowicz, Ichter, Driess, Nair, Vuong,
  Mees, Finn, and Levine]{pertsch2025fast}
Karl Pertsch, Kyle Stachowicz, Brian Ichter, Danny Driess, Suraj Nair, Quan
  Vuong, Oier Mees, Chelsea Finn, and Sergey Levine.
\newblock Fast: Efficient action tokenization for vision-language-action
  models.
\newblock \emph{arXiv preprint arXiv:2501.09747}, 2025.

\bibitem[Qin et~al.(2020)Qin, Fang, Zhu, Fei-Fei, and Savarese]{qin2020keto}
Zengyi Qin, Kuan Fang, Yuke Zhu, Li~Fei-Fei, and Silvio Savarese.
\newblock Keto: Learning keypoint representations for tool manipulation.
\newblock In \emph{2020 IEEE International Conference on Robotics and
  Automation (ICRA)}, pages 7278--7285. IEEE, 2020.

\bibitem[Raffel et~al.(2020)Raffel, Shazeer, Roberts, Lee, Narang, Matena,
  Zhou, Li, and Liu]{Raffel2020ExploringTL}
Colin Raffel, Noam Shazeer, Adam Roberts, Katherine Lee, Sharan Narang, Michael
  Matena, Yanqi Zhou, Wei Li, and Peter~J. Liu.
\newblock Exploring the limits of transfer learning with a unified text-to-text
  transformer.
\newblock \emph{Journal of Machine Learning Research}, 21:\penalty0 1--67,
  2020.
\newblock URL \url{https://jmlr.org/papers/v21/20-074.html}.

\bibitem[Ray et~al.(2024)Ray, Duan, Tan, Bashkirova, Hendrix, Ehsani, Kembhavi,
  Plummer, Krishna, Zeng, et~al.]{ray2024sat}
Arijit Ray, Jiafei Duan, Reuben Tan, Dina Bashkirova, Rose Hendrix, Kiana
  Ehsani, Aniruddha Kembhavi, Bryan~A Plummer, Ranjay Krishna, Kuo-Hao Zeng,
  et~al.
\newblock Sat: Spatial aptitude training for multimodal language models.
\newblock \emph{arXiv preprint arXiv:2412.07755}, 2024.

\bibitem[Ren et~al.(2024)Ren, Liu, Zeng, Lin, Li, Cao, Chen, Huang, Chen, Yan,
  et~al.]{ren2024grounded}
Tianhe Ren, Shilong Liu, Ailing Zeng, Jing Lin, Kunchang Li, He~Cao, Jiayu
  Chen, Xinyu Huang, Yukang Chen, Feng Yan, et~al.
\newblock Grounded sam: Assembling open-world models for diverse visual tasks.
\newblock \emph{arXiv preprint arXiv:2401.14159}, 2024.

\bibitem[Shao et~al.(2024)Shao, Qian, Xiao, Song, Zong, Wang, Liu, and
  Li]{shao2024visual}
Hao Shao, Shengju Qian, Han Xiao, Guanglu Song, Zhuofan Zong, Letian Wang,
  Yu~Liu, and Hongsheng Li.
\newblock Visual cot: Advancing multi-modal language models with a
  comprehensive dataset and benchmark for chain-of-thought reasoning.
\newblock In \emph{The Thirty-eight Conference on Neural Information Processing
  Systems Datasets and Benchmarks Track}, 2024.

\bibitem[Shridhar et~al.(2024)Shridhar, Lo, and James]{shridhar2024generative}
Mohit Shridhar, Yat~Long Lo, and Stephen James.
\newblock Generative image as action models.
\newblock \emph{arXiv preprint arXiv:2407.07875}, 2024.

\bibitem[Song et~al.(2024)Song, Blukis, Tremblay, Tyree, Su, and
  Birchfield]{song2024robospatial}
Chan~Hee Song, Valts Blukis, Jonathan Tremblay, Stephen Tyree, Yu~Su, and Stan
  Birchfield.
\newblock Robospatial: Teaching spatial understanding to 2d and 3d
  vision-language models for robotics.
\newblock \emph{arXiv preprint arXiv:2411.16537}, 2024.

\bibitem[Sundaralingam et~al.(2023)Sundaralingam, Hari, Fishman, Garrett,
  Van~Wyk, Blukis, Millane, Oleynikova, Handa, Ramos,
  et~al.]{sundaralingam2023curobo}
Balakumar Sundaralingam, Siva Kumar~Sastry Hari, Adam Fishman, Caelan Garrett,
  Karl Van~Wyk, Valts Blukis, Alexander Millane, Helen Oleynikova, Ankur Handa,
  Fabio Ramos, et~al.
\newblock Curobo: Parallelized collision-free robot motion generation.
\newblock In \emph{2023 IEEE International Conference on Robotics and
  Automation (ICRA)}, pages 8112--8119. IEEE, 2023.

\bibitem[Team et~al.(2024)Team, Team, Brohan, Brown, Chen, Cheng, Choromanski,
  Cullina, Dalal, Fu, Golemo, et~al.]{team2024octo}
Octo Team, RT-X Team, Anthony Brohan, Noah Brown, Lauren Chen, Michael Cheng,
  Krzysztof Choromanski, Eamonn Cullina, Gabe Dalal, Chelsea Fu, Florian
  Golemo, et~al.
\newblock Octo: An open-source generalist robot policy.
\newblock \emph{arXiv preprint arXiv:2403.10164}, 2024.

\bibitem[Tong et~al.(2024)Tong, Brown, Wu, Woo, Middepogu, Akula, Yang, Yang,
  Iyer, Pan, Wang, Fergus, LeCun, and Xie]{tong2024cambrian1}
Shengbang Tong, Ellis Brown, Penghao Wu, Sanghyun Woo, Manoj Middepogu,
  Sai~Charitha Akula, Jihan Yang, Shusheng Yang, Adithya Iyer, Xichen Pan,
  Austin Wang, Rob Fergus, Yann LeCun, and Saining Xie.
\newblock Cambrian-1: A fully open, vision-centric exploration of multimodal
  llms, 2024.

\bibitem[Vaswani et~al.(2017)Vaswani, Shazeer, Parmar, Uszkoreit, Jones, Gomez,
  Kaiser, and Polosukhin]{vaswani2017attention}
Ashish Vaswani, Noam Shazeer, Niki Parmar, Jakob Uszkoreit, Llion Jones,
  Aidan~N Gomez, {\L}ukasz Kaiser, and Illia Polosukhin.
\newblock Attention is all you need.
\newblock \emph{Advances in neural information processing systems}, 30, 2017.

\bibitem[Walke et~al.(2023)Walke, Black, Zhao, Vuong, Zheng, Hansen-Estruch,
  He, Myers, Kim, Du, et~al.]{walke2023bridgedata}
Homer~Rich Walke, Kevin Black, Tony~Z Zhao, Quan Vuong, Chongyi Zheng, Philippe
  Hansen-Estruch, Andre~Wang He, Vivek Myers, Moo~Jin Kim, Max Du, et~al.
\newblock Bridgedata v2: A dataset for robot learning at scale.
\newblock In \emph{Conference on Robot Learning}, pages 1723--1736. PMLR, 2023.

\bibitem[Wang et~al.(2024)Wang, Chen, Zhao, and He]{wang2024scaling}
Lirui Wang, Xinlei Chen, Jialiang Zhao, and Kaiming He.
\newblock Scaling proprioceptive-visual learning with heterogeneous pre-trained
  transformers.
\newblock \emph{arXiv preprint arXiv:2409.20537}, 2024.

\bibitem[Wang et~al.(2025)Wang, Ren, Luo, Li, Yan, Chen, Wang, Li, Lu, Zhu,
  et~al.]{wang2025all}
Weiyun Wang, Yiming Ren, Haowen Luo, Tiantong Li, Chenxiang Yan, Zhe Chen,
  Wenhai Wang, Qingyun Li, Lewei Lu, Xizhou Zhu, et~al.
\newblock The all-seeing project v2: Towards general relation comprehension of
  the open world.
\newblock In \emph{European Conference on Computer Vision}, pages 471--490.
  Springer, 2025.

\bibitem[Wei et~al.(2022)Wei, Wang, Schuurmans, Bosma, Xia, Chi, Le, Zhou,
  et~al.]{wei2022chain}
Jason Wei, Xuezhi Wang, Dale Schuurmans, Maarten Bosma, Fei Xia, Ed~Chi, Quoc~V
  Le, Denny Zhou, et~al.
\newblock Chain-of-thought prompting elicits reasoning in large language
  models.
\newblock \emph{Advances in neural information processing systems},
  35:\penalty0 24824--24837, 2022.

\bibitem[Wen et~al.(2023)Wen, Lin, So, Chen, Dou, Gao, and Abbeel]{wen2023any}
Chuan Wen, Xingyu Lin, John So, Kai Chen, Qi~Dou, Yang Gao, and Pieter Abbeel.
\newblock Any-point trajectory modeling for policy learning.
\newblock \emph{arXiv preprint arXiv:2401.00025}, 2023.

\bibitem[Wu et~al.(2025)Wu, Wang, Tang, Wu, He, Ouyang, Torr, and
  Wu]{wu2025dettoolchain}
Yixuan Wu, Yizhou Wang, Shixiang Tang, Wenhao Wu, Tong He, Wanli Ouyang, Philip
  Torr, and Jian Wu.
\newblock Dettoolchain: A new prompting paradigm to unleash detection ability
  of mllm.
\newblock In \emph{European Conference on Computer Vision}, pages 164--182.
  Springer, 2025.

\bibitem[Xu et~al.(2024)Xu, Xu, Xu, Chi, Wetzstein, Veloso, and
  Song]{xu2024flow}
Mengda Xu, Zhenjia Xu, Yinghao Xu, Cheng Chi, Gordon Wetzstein, Manuela Veloso,
  and Shuran Song.
\newblock Flow as the cross-domain manipulation interface.
\newblock \emph{arXiv preprint arXiv:2407.15208}, 2024.

\bibitem[Yang et~al.(2025)Yang, Tan, Wu, Zheng, Peng, Liang, Gu, Cai, Ye, Jang,
  et~al.]{yang2025magma}
Jianwei Yang, Reuben Tan, Qianhui Wu, Ruijie Zheng, Baolin Peng, Yongyuan
  Liang, Yu~Gu, Mu~Cai, Seonghyeon Ye, Joel Jang, et~al.
\newblock Magma: A foundation model for multimodal ai agents.
\newblock \emph{arXiv preprint arXiv:2502.13130}, 2025.

\bibitem[Yao et~al.(2023{\natexlab{a}})Yao, Tian, Liu, Zhang, Liu, Jin, Li, Li,
  and Sun]{yao2023thinking}
Fanglong Yao, Changyuan Tian, Jintao Liu, Zequn Zhang, Qing Liu, Li~Jin,
  Shuchao Li, Xiaoyu Li, and Xian Sun.
\newblock Thinking like an expert: Multimodal hypergraph-of-thought (hot)
  reasoning to boost foundation modals.
\newblock \emph{arXiv preprint arXiv:2308.06207}, 2023{\natexlab{a}}.

\bibitem[Yao et~al.(2024)Yao, Yu, Zhao, Shafran, Griffiths, Cao, and
  Narasimhan]{yao2024tree}
Shunyu Yao, Dian Yu, Jeffrey Zhao, Izhak Shafran, Tom Griffiths, Yuan Cao, and
  Karthik Narasimhan.
\newblock Tree of thoughts: Deliberate problem solving with large language
  models.
\newblock \emph{Advances in Neural Information Processing Systems}, 36, 2024.

\bibitem[Yao et~al.(2023{\natexlab{b}})Yao, Li, and Zhao]{yao2023beyond}
Yao Yao, Zuchao Li, and Hai Zhao.
\newblock Beyond chain-of-thought, effective graph-of-thought reasoning in
  language models.
\newblock \emph{arXiv preprint arXiv:2305.16582}, 2023{\natexlab{b}}.

\bibitem[You et~al.(2023)You, Zhang, Gan, Du, Zhang, Wang, Cao, Chang, and
  Yang]{you2023ferret}
Haoxuan You, Haotian Zhang, Zhe Gan, Xianzhi Du, Bowen Zhang, Zirui Wang,
  Liangliang Cao, Shih-Fu Chang, and Yinfei Yang.
\newblock Ferret: Refer and ground anything anywhere at any granularity.
\newblock \emph{arXiv preprint arXiv:2310.07704}, 2023.

\bibitem[Yuan et~al.(2024{\natexlab{a}})Yuan, Wen, Zhang, and
  Gao]{yuan2024general}
Chengbo Yuan, Chuan Wen, Tong Zhang, and Yang Gao.
\newblock General flow as foundation affordance for scalable robot learning.
\newblock \emph{arXiv preprint arXiv:2401.11439}, 2024{\natexlab{a}}.

\bibitem[Yuan et~al.(2024{\natexlab{b}})Yuan, Duan, Blukis, Pumacay, Krishna,
  Murali, Mousavian, and Fox]{yuan2024robopoint}
Wentao Yuan, Jiafei Duan, Valts Blukis, Wilbert Pumacay, Ranjay Krishna,
  Adithyavairavan Murali, Arsalan Mousavian, and Dieter Fox.
\newblock Robopoint: A vision-language model for spatial affordance prediction
  for robotics.
\newblock \emph{arXiv preprint arXiv:2406.10721}, 2024{\natexlab{b}}.

\bibitem[Zawalski et~al.(2024)Zawalski, Chen, Pertsch, Mees, Finn, and
  Levine]{zawalski2024robotic}
Micha{\l} Zawalski, William Chen, Karl Pertsch, Oier Mees, Chelsea Finn, and
  Sergey Levine.
\newblock Robotic control via embodied chain-of-thought reasoning.
\newblock \emph{arXiv preprint arXiv:2407.08693}, 2024.

\bibitem[Zhang et~al.(2024)Zhang, Li, Li, Zeng, Zhao, Sun, Chen, Wei, Zhan, Li,
  et~al.]{zhang2024empowering}
Tianle Zhang, Dongjiang Li, Yihang Li, Zecui Zeng, Lin Zhao, Lei Sun, Yue Chen,
  Xuelong Wei, Yibing Zhan, Lusong Li, et~al.
\newblock Empowering embodied manipulation: A bimanual-mobile robot
  manipulation dataset for household tasks.
\newblock \emph{arXiv preprint arXiv:2405.18860}, 2024.

\bibitem[Zhang et~al.(2022)Zhang, Zhang, Li, and Smola]{zhang2022automatic}
Zhuosheng Zhang, Aston Zhang, Mu~Li, and Alex Smola.
\newblock Automatic chain of thought prompting in large language models.
\newblock \emph{arXiv preprint arXiv:2210.03493}, 2022.

\bibitem[Zheng et~al.(2023{\natexlab{a}})Zheng, Yang, Tang, Zhou, and
  Yang]{zheng2023ddcot}
Ge~Zheng, Bin Yang, Jiajin Tang, Hong-Yu Zhou, and Sibei Yang.
\newblock Ddcot: Duty-distinct chain-of-thought prompting for multimodal
  reasoning in language models.
\newblock \emph{Advances in Neural Information Processing Systems},
  36:\penalty0 5168--5191, 2023{\natexlab{a}}.

\bibitem[Zheng et~al.(2023{\natexlab{b}})Zheng, Chiang, Sheng, Zhuang, Wu,
  Zhuang, Lin, Li, Li, Xing, et~al.]{zheng2023judging}
Lianmin Zheng, Wei-Lin Chiang, Ying Sheng, Siyuan Zhuang, Zhanghao Wu, Yonghao
  Zhuang, Zi~Lin, Zhuohan Li, Dacheng Li, Eric Xing, et~al.
\newblock Judging llm-as-a-judge with mt-bench and chatbot arena.
\newblock \emph{Advances in Neural Information Processing Systems},
  36:\penalty0 46595--46623, 2023{\natexlab{b}}.

\bibitem[Zheng et~al.(2024)Zheng, Liang, Huang, Gao, Daum{\'e}~III, Kolobov,
  Huang, and Yang]{zheng2024tracevla}
Ruijie Zheng, Yongyuan Liang, Shuaiyi Huang, Jianfeng Gao, Hal Daum{\'e}~III,
  Andrey Kolobov, Furong Huang, and Jianwei Yang.
\newblock Tracevla: Visual trace prompting enhances spatial-temporal awareness
  for generalist robotic policies.
\newblock \emph{arXiv preprint arXiv:2412.10345}, 2024.

\bibitem[Zhu et~al.(2024)Zhu, Kumar, Hu, and Liu]{zhu2024tame}
Shengjie Zhu, Abhinav Kumar, Masa Hu, and Xiaoming Liu.
\newblock Tame a wild camera: in-the-wild monocular camera calibration.
\newblock \emph{Advances in Neural Information Processing Systems}, 36, 2024.

\end{thebibliography}

\appendix
\newpage

\appendix

\section*{\hspace{-4mm} \centering Appendix}
\vspace{3mm}

\section{Weak-to-Strong Dataset Construction}
\label{app:weak-to-strong dataset}

\textbf{Region Grounding Dataset Generation} Regional grounding enables robots to focus on key objects within a scene. In embodied AI, robots need to concentrate on task-relevant local scenes in the image according to task instructions and be able to accurately locate objects. In Level 1 of our dataset, we integrate grounding with captioning, aiming for the agent to provide positional information when describing images. Unlike traditional image captioning, objects in embodied scenes are often cluttered, and we only seek to extract captions for task-relevant regions, avoiding redundant information. We prompt GPT-4o to exclude task-irrelevant regions based on task instructions and generate both image captions and the corresponding object names. Subsequently, we use Grounding DINO~\cite{liu2023grounding} to capture the locations of various objects in the image and embed them into the caption sentences, forming captions with object location information. 
Below is an example from the dataset:

\begin{tcolorbox}
[colback=orange!5!white,
colframe=orange!80!black, title={Level 1: Region Grounding}]

\texttt{\textbf{Human:} Give a short and clear explanation of the subsequent image.
\textbf{GPT:} A <ref>robotic arm</ref><box>[[338, 126, 994, 861]]</box> is positioned above a <ref>blue bottle</ref><box>[[257, 201, 381, 413]]</box>, with a <ref>grey toy</ref><box>[[391, 413, 538, 518]]</box> to the left and a <ref>green oval object</ref><box>[[592, 481, 702, 601]]</box> to the right on a wooden table.}
\end{tcolorbox}

\textbf{Spatial Relationship Dataset Generation} To accurately infer object spatial relationships from RGB images, a multi-stage pipeline is employed: encompassing object detection, instance segmentation, 2D-to-3D mapping, and subsequent relationship calculation.

\begin{figure}[ht]
    \centering
    \includegraphics[width=1\linewidth]{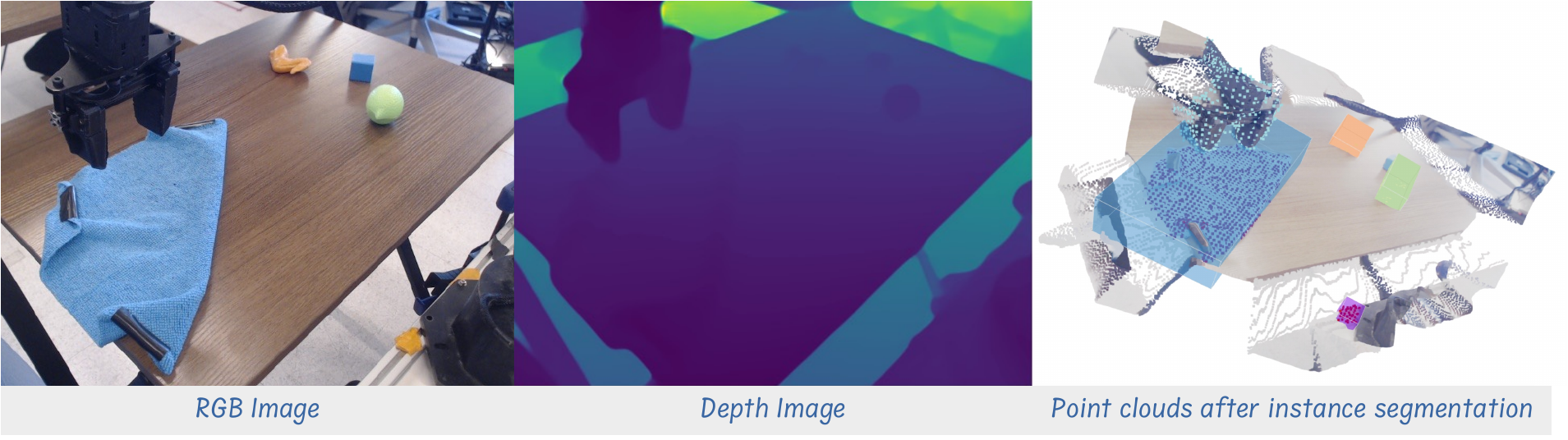}
    \caption{RGB-D images before transformation and the visualized point clouds after transformation.}
    \label{fig:vis_pcd}
\end{figure}

Initially, for pre-identified objects of interest within the scene, GroundedSAM~\cite{ren2024grounded} is utilized to perform instance segmentation, yielding precise object masks. Subsequently, the 2D RGB information is elevated to a 3D spatial representation. This transformation begins with leveraging PerspectiveFields~\cite{jin2023perspective} to estimate the Z-axis orientation, serving as a coarse approximation for camera extrinsics. Concurrently, the WildCamera~\cite{zhu2024tame} model is employed to estimate intrinsic camera parameters, including focal length and resolution. Metric3Dv2~\cite{hu2024metric3d} is then used for robust depth estimation. By integrating the RGB image, estimated depth image, and computed camera intrinsics and extrinsics, the 2D RGB image is transformed into a 3D point cloud. Given the prior object segmentation, the specific 3D spatial position and size of each individual object can be precisely derived from the generated point cloud. These comprehensive 3D data then enable the calculation of relative spatial relationships between objects, which are subsequently exported as a spatial relationship graph. We present the 2D RGB-D images before transformation and the visualized point clouds after transformation in \cref{fig:vis_pcd}.

It is important to note that for inferring spatial relationships, only relative depth sorting data is generated, which means stringent accuracy requirements for absolute depth are relaxed. To further enhance data quality, particularly for robust relationship inference, objects exhibiting a relative depth gap of at least 20\% are preferentially selected for subsequent generation. An example from the dataset: 

\begin{tcolorbox}[colback=orange!5!white,
colframe=orange!80!black, title={Level 2: Spatial Relationship}]

\texttt{Human: Tell me about the items in your view and their relationships to each other. Answer the question with scene graph. \\
GPT: <ref>front left burner</ref><box>[[597, 422, 817, 596]]</box> is positioned <pred>to the right of</pred><box>[[597, 422, 817, 596]]</box><box>[[171, 275, 446, 409]]</box> <ref>metal pot</ref><box>[[171, 275, 446, 409]]</box>. <ref>metal pot</ref><box>[[171, 275, 446, 409]]</box> appears <pred>to the left of</pred><box>[[171, 275, 446, 409]]</box><box>[[431, 474, 715, 608]]</box> <ref>green spatula</ref><box>[[431, 474, 715, 608]]</box>. <ref>metal pot</ref><box>[[171, 275, 446, 409]]</box> can be seen <pred>to the left of</pred><box>[[171, 275, 446, 409]]</box><box>[[519, 408, 600, 490]]</box> <ref>orange cheese slice</ref><box>[[519, 408, 600, 490]]</box>.}
\end{tcolorbox}

\textbf{Spatial Reasoning QA Generation} After generating the spatial relation diagrams, we can easily create various template questions for spatial reasoning~\cite{cheng2024spatialrgpt, du2024embspatial}, such as: \textit{How are [A] and [B] positioned in relation to each other in the image? or From your perspective, which object in the image is at the shortest distance?} In addition to template-based QA, we also combine task instructions, existing spatial information, and images to query GPT-4o, thereby generating more diverse multi-turn dialogues to enhance the model’s generalization ability in spatial reasoning.

\textbf{Spatial Affordance and Visual Trace Dataset Generation} Next, we provide a detailed description of how to extract the required controllable points/boxes and visual trajectories from the embodied dataset like BridgeData~\cite{walke2023bridgedata}, RT-X~\cite{o2023open} and Droid~\cite{khazatsky2024droid}. Our methodology for extracting visual aids, specifically Affordance Boxes, Affordance Points, and Visual Traces, involves a multi-stage process leveraging state-of-the-art vision models. We also incorporate a rigorous data validation procedure to ensure high-quality output.

First, we acquire the initial and final frames of the video sequence. To determine the Affordance Box, we utilize GroundingDINO~\cite{liu2023grounding} and GroundedSam~\cite{ren2024grounded} to detect the mask of the \texttt{manipulated\_object} in the final frame. The bounding box of this detected object in the final frame defines the Affordance Box, representing the ultimate spatial location of the manipulated object. Subsequently, we extract Affordance Points. This involves performing an erosion operation on the mask of the \texttt{manipulated\_object} obtained from the final frame. Erosion reduces the mask's area, facilitating the sampling of points that are more central or internal to the object, thereby mitigating the risk of sampling edge points. From this eroded mask, we uniformly sample 8 points, which constitute the Affordance Points. For Visual Trace extraction, we begin by detecting the mask of the \texttt{manipulated\_object} in the first frame. From this mask, we sample 3 points, which serve as the initial query points for the CoTracker~\cite{karaev2024cotracker3} model. Selecting multiple points enhances the robustness of trajectory tracking. The processed video sequence and these sampled points are then fed into the CoTracker model, which outputs the predicted trajectory for each query point across every frame of the video sequence. We calculate the total distance of each trajectory and select the longest trajectory as the representative trajectory for the \texttt{manipulated\_object}. This chosen trajectory is then subjected to cubic spline interpolation for smoothing. Finally, 8 equidistant points are uniformly sampled from the smoothed trajectory, forming the Visual Trace.

A critical aspect of this process is addressing potential prediction errors from the pre-trained visual models, such as incorrect object identification or incomplete tracking of object motion. To mitigate these issues, we implement stringent rule-based filtering using hyperparameters like size thresholds and trajectory length thresholds. Before annotating each dataset, we iteratively adjust these rules and conduct manual inspections on 100 examples. Only when the accuracy of these filtered results surpasses 95\% do we deem the filtering rules robust enough to proceed with the full data generation pipeline. This meticulous validation process ensures the high quality of the generated data.

Following the generation of these visual aids, we also pre-generate the thinking processes for Chain-of-Thought (CoT) reasoning. We input templates, questions, and answers into GPT-4o, querying it to complete the thinking process. The complete query prompt is constructed accordingly: 

\begin{tcolorbox}[title={Prompt and In-context Example for Completing Thinking Process}]

\texttt{You are an AI visual assistant that can analyze a single image. You receive one image
and corresponding caption, task instruction, manipulated object in the task and target place to finish the task,
and bounding box position and relative position of these objects. Also, you receive the answer of points and bbox.
Now, you need to generate a <Description>...</Description>, <Reasoning>...</Reasoning>, <Answer>...</Answer> format.}

\texttt{<Description>
First, using the provided caption, task instruction, describe the scene.
If there are errors in the caption, please ignore them and do not point them out in your description.
Instead of directly mentioning the bounding box coordinates, utilize this data to explain the scene using natural language
with its bounding box in the format like "<ref>object</ref><box>[[x1, y1, x2, y2]]</box>".
When mentioning the predicate between two objects, you should mention it in the format like "<pred>predicate</pred><box>[[x1, y1, x2, y2]]</box><box>[[x3, y3, x4, y4]]</box>", where "<box>[[x1, y1, x2, y2]]</box>" denotes
the bounding box coordinates of the subject and "<box>[[x3, y3, x4, y4]]</box>" denotes the bounding box coordinates of the object.
</Description>}

\texttt{<Reasoning>
According to the task instruction and the answer of points and bbox, provide a chain-of-thought, logical explanation of the problem. 
</Reasoning>}

\texttt{<Answer>
State the final bounding box answer and point answer in a clear and direct format here. 
Bounding box answer is in the format like "<box>[[x1, y1, x2, y2]]</box>".
Point answer is in the format like "<point>[[x1, y1], [x2, y2], [x3, y3], ...]</point>".
</Answer>}

\texttt{Here is an example: \\
Caption: 7up can is on the table beside the bowl, drawer is open.\\
Task instruction: place 7up can into top drawer \\
Manipulated object: 7up can<box>[[686, 174, 788, 357]]</box> \\
Target place: top drawer <box>[[135, 411, 852, 920]]</box>
Current Relation: (7up can, [686, 174, 788, 357], top drawer, [135, 411, 852, 920], above) \\
Bounding Box Answer: [569, 553, 651, 679] \\
Point Answer: [[603, 609], [593, 637], [596, 643], [615, 596], [612, 593], [590, 637], [603, 606], [615, 631]] \\
<Description>
The scene describes a <ref>7up can</ref><box>[[686, 174, 788, 357]]</box> that is positioned on a table. 
It is situated <pred>above</pred><box>[[686, 174, 788, 357]]</box><box>[[135, 411, 852, 920]]</box> the
<ref>top drawer</ref><box>[[135, 411, 852, 920]]</box> which is open. 
</Description>
<Reasoning>
The 7up can is above the top drawer now and should be placed into the top drawer. So the target position of the 7up
can might be similar to the position of the drawer's interior.
</Reasoning>
<Answer>
To complete the task, the target position is <box>[[569, 553, 651, 679]]</box>.
The new position can also be roughly defined by the following several free points 
<point>[[603, 609], [593, 637], [596, 643], [615, 596], [612, 593], [590, 637], [603, 606], [615, 631]]</point>,
 which can outline the position of the 7up can after it is placed in the top drawer.
</Answer>
}

\end{tcolorbox}

Next, we present two examples from the dataset: one involves generating Affordance Boxes/Points (Level 4), and the other involves generating visual trace (Level 5). We consider Affordance Points as samples within the box; thus, generating the box first and then the points is a refinement process from coarse to fine granularity. Therefore, both types are generated for each data, and one can choose which to use as needed. In the task of generating visual trace, since we performed equidistant interpolation in advance, the visual trace in the dataset is fixed at 8 points. Consequently, models trained on such a dataset will also generate 8 points as the visual trace.

\begin{tcolorbox}[colback=orange!5!white,
colframe=orange!80!black, title={Level 4: Spatial Affordance Box/Points Generation}]

\texttt{Human: "You are currently a robot performing robotic manipulation tasks. Your task instruction: Moved the can to the left side of the green cloth. Observe the image, use 2D points and bounding box to mark the target location where the manipulated object will be moved. In your answer, use <box>[[x1, y1, x2, y2]]</box> to present the bounding box of the target region, and use <point>[[x1, y1], [x2, y2], ...]</point> to mark the points of the free space."} \\

\texttt{GPT: "<Description>\\The image shows a <ref>white soup can</ref><box>[[544, 177, 700, 368]]</box> positioned to the right of a <ref>green towel</ref><box>[[244, 389, 664, 676]]</box> on a table. A small bowl containing a yellow object and a brush is next to the towel.\\</Description>\\<Reasoning>\\The task is to move the white soup can to the left side of the green towel. Currently, the can is to the right of the towel, so it needs to be repositioned to the left of the towel's current location.\\</Reasoning>\\<Answer>\\The target position for the white soup can is <box>[[250, 181, 400, 392]]</box>. The new position can also be roughly defined by the following points: <point>[[346, 248], [302, 365], [377, 251], [330, 295], [357, 291], [354, 362], [329, 355], [312, 352]]</point>.\\</Answer>"}
\end{tcolorbox}

\begin{tcolorbox}[colback=orange!5!white,
colframe=orange!80!black, title={Level 5: Visual Trace Generation}]

\texttt{Human: "You are currently a robot performing robotic manipulation tasks. Your task instruction: put the green spatula in the silver pot. Observe the image, use 2D points to mark the manipulated object-centric waypoints to guide the robot to manipulate the object. Typically, the waypoints consist of an ordered sequence of eight 2D points. The format is <point>[[x1, y1], [x2, y2], ...]</point>."} \\

\texttt{GPT: "<Description>\\The image shows a <ref>green spatula</ref><box>[[762, 536, 856, 711]]</box> positioned to the right of an orange towel on a table. A <ref>silver pot</ref><box>[[0, 723, 242, 872]]</box> is located above a robotic arm. Other objects, such as a gray can and a mushroom, are also present in the scene.\\</Description>\\<Reasoning>\\To move the green spatula into the silver pot, start by identifying the initial position of the spatula at <point>[[802, 613]]</point>. First, lift the spatula upwards to clear the table, moving to <point>[[774, 501]]</point>. Then, move it horizontally towards the left, passing through <point>[[685, 394]]</point>. Finally, lower the spatula into the silver pot, ending at <point>[[657, 401]]</point> within the target bounding box <box>[[225, 529, 407, 640]]</box>.\\</Reasoning>\\<Answer>\\The visual trace for placing the green spatula into the silver pot is <point>[[802, 613], [780, 582], [774, 501], [744, 465], [685, 394], [657, 349], [668, 354], [657, 401]]</point>.\\</Answer>"} \\

\end{tcolorbox}

\section{Training Details and Datasets}
\label{app:training details}

The training of \alg is a two-stage process, building upon the LLaVA~\cite{liu2024llavanext} architecture through continued fine-tuning of ASMv2~\cite{wang2025all}. In order to train a VLM with powerful spatiotemporal reasoning abilities that can also generate visual aids, we leveraged a comprehensive dataset of approximately 1.4M samples from various sources. \textit{Note:} The position coordinate format used by FSD normalizes the image coordinates to a range of 0-999 after padding the image to a square shape. For all datasets mentioned below, the coordinates have been pre-processed in this manner.

\textbf{Stage 1: Spatial Reasoning Enhancement} The primary goal of the first stage is to enhance the model's spatial reasoning capabilities.  This dataset is primarily composed of two parts: common-sense image QA and conversations data and spatial reasoning data. The inclusion of GeneralQA is crucial for FSD to maintain broad instruction-following abilities after fine-tuning.
First, we selected some datasets and randomly sampled around 838k data from LLaVA-665k and ASMv2-4M instruction-following data. Next, we incorporated approximately 295k samples of spatial reasoning data from the LLaVA-OneVision~\cite{li2024llava}, RoboPoint~\cite{yuan2024robopoint}, SpatialBot~\cite{cai2024spatialbot} and SAT~\cite{ray2024sat} training datasets. Finally, the first three levels of FSD's collected embodied spatial reasoning datasets~(250k samples) were also integrated into this training phase. A summary of the datasets used can be found in \cref{tab:fsd_dataset}.

\textbf{Stage 2: Visual Aids Generation and Understanding} We specifically focused on training for visual aids generation. Here, "generation" refers to generating visual aids based on a given question, while "understanding" represents its inverse problem. We excluded some inverse problems with ambiguous semantics. The final total amount of data used is shown in \cref{tab:fsd_dataset}.

\textbf{Training Configuration} We used exactly the same hyperparameters in both stages of training. We utilized a global batch size of 128 and the AdamW optimizer, configured with $\beta_1 = 0.9$, $\beta_2 = 0.999$, and a weight decay coefficient of 0. The learning rate was set to $2 \times 10^{-5}$. Both training stages employ a linear warmup for the first 3\% of training steps, followed by a cosine decay strategy to a minimum learning rate of 0. We simultaneously train both the vision-language connector and the language model. The image resolution was set to $336 \times 336$, and the visual encoder remained frozen throughout the entire training process. In the first phase, we train for 1 epoch on the complete dataset, while in the second phase, we train for 3 epochs. The \alg model has approximately 13B trainable parameters and we conducted training using 8 A100 40G GPUs, with Stage 1 requiring approximately 72 hours and Stage 2 requiring 8 hours.

\section{Details of Action Execution}
\label{app:action execution}

When utilizing FSD for robotic manipulation tasks, we can select from various visual aids. As described in \cref{fig:visual_aids}, these include \textbf{spatial affordance boxes} ($\mathcal{B}$), \textbf{spatial affordance points} ($\mathcal{P}$), and \textbf{object-centric visual traces} ($\boldsymbol{\tau}$). The choice of visual aids dictates the subsequent motion planning strategy.

\textbf{Motion Planning with Spatial Affordances} For spatial affordance boxes ($\mathcal{B}$), the target point for manipulation is derived by sampling the center of the box. In the case of spatial affordance points ($\mathcal{P}$), a point is directly sampled. When relying on spatial affordance information, whether from boxes or points, we employ CuRobo~\cite{sundaralingam2023curobo} as our motion planner. CuRobo is responsible for generating collision-free paths that guide the robot's end-effector to the inferred target affordance point.

\textbf{Motion Planning with Object-Centric Visual Traces} When leveraging object-centric visual traces ($\boldsymbol{\tau}$), the process involves mapping 2D visual traces into 3D space and then interpolating these discrete points to form a complete motion trajectory in SE(3) space. The detailed procedure is as follows:

We directly acquire 2D keypoint information, denoted as $k_i=(u_i,v_i)\in \mathbb{R}^2$, where $u_i$ and $v_i$ represent the x and y coordinates of the $i$-th point in the image, for $i \in [1,T]$. Initial depth information, $d_i\in \mathbb{R}$, is obtained from a depth camera. Using the Pinhole camera model~\cite{HartleyZisserman2004}, we can transform these 2D keypoints into 3D Cartesian coordinates $P_i=(x_i,y_i,z_i)\in \mathbb{R}^3$ via:

$$s_i \begin{bmatrix} u_i \\ v_i \\ 1 \end{bmatrix} = \begin{bmatrix} f_x & 0 & c_x \\ 0 & f_y & c_y \\ 0 & 0 & 1 \end{bmatrix} \begin{bmatrix} x_i \\ y_i \\ z_i \end{bmatrix}$$

Here, $s_i$ is the normalized depth, calculated as $s_i=d_i/\texttt{depth\_scale}$. The intrinsic camera parameters, $f_x, f_y, c_x, c_y$, and the depth scaling factor, $\texttt{depth\_scale}$, are all camera-specific.

However, a naive use of the raw depth values $d_i$ obtained from the depth camera often results in trajectories that closely hug the object's surface, which is undesirable for robust robot manipulation. To address this, we formulate an optimization problem to address it. We fix the depth values for the start and end points of the path ($d_1$ and $d_T$) and optimize the intermediate depth values $d_{2:T-1}$ to minimize the total Euclidean distance between consecutive points in Cartesian space.

The objective function for this optimization is:

$$\hat d_i = \arg\min_{d_{2:T-1}}\sum_i{d(P_i,P_{i+1})}$$

where $d(P_i,P_{i+1})$ represents the Euclidean distance between points $P_i$ and $P_{i+1}$. We employ a gradient descent method from \texttt{scipy} library to optimize this objective function. This refined approach allows for more robust and practical robot motion planning by addressing the limitations of raw depth data and providing a structured framework for integrating various visual aids.

\section{Details of Visual Aids Generation Benchmark}
\label{app:vabench}

To address the gap in evaluating visual assistance signal generation, we established \textbf{VABench}, a comprehensive benchmark. VABench comprises two distinct tasks: \textbf{VABench-Point} and \textbf{VABench-VisualTrace}, each featuring 300 meticulously hand-annotated questions. These tasks require models to infer visual auxiliary information solely from natural language instructions, mimicking everyday human commands.

For \textbf{VABench-Point}, we provide ground truth bounding boxes for each question. The model's performance is then assessed by calculating the proportion of predicted points that fall within the target region. For models that only output bounding boxes, we explored two scoring methods: Intersection Over Union (IOU) and uniformly sampling points within the predicted box. We ultimately opted for the latter approach, as the physical interaction between a robotic arm and an object in real-world tasks is fundamentally determined by specific point coordinates. For \textbf{VABench-VisualTrace}, we provide ground truth trajectories consisting of eight points. When the predicted trajectory length deviates from the ground truth, we employ interpolation to align their lengths. To ensure consistent evaluation across varying image resolutions, all coordinates are uniformly normalized to a range of $0$ to $1000$. Subsequently, we employ a combination of Mean Absolute Error (MAE), Root Mean Squared Error (RMSE), and \textbf{GPT Score} to comprehensively evaluate performance.

Here, we detail the design philosophy and evaluation procedure of the GPT Score. Given that multiple valid solutions can exist for each task instruction, relying solely on trajectory similarity to the ground truth as a scoring criterion can be one-sided. Therefore, we established detailed evaluation criteria to simulate realistic human assessment. Based on these rules, we introduced a visualized trajectory scoring method leveraging Multimodal Large Language Models (MLLMs), termed \textbf{GPT-score}.

Specifically, we designed an evaluation prompt to guide GPT-4.1 in assessing predicted object manipulation trajectories based on both task instructions and visual inputs. The prompt provides clear instructions and criteria, positioning the model as an expert evaluator in robotic manipulation and visual reasoning. Each evaluation instance consists of a task instruction and an accompanying image that visualizes the predicted trajectory, where a red circle indicates the start point and a blue diamond marks the end point.
The model is instructed to assess the trajectory according to three key criteria: (1) task alignment and success, determining whether the predicted path correctly fulfills the instruction by starting and ending in appropriate locations; (2) feasibility, evaluating the physical plausibility and smoothness of the motion; and (3) obstacle avoidance, considering whether the trajectory avoids potential collisions. The prompt emphasizes that completing the task correctly is the most important factor; any major deviation in goal achievement results in a low score, even if the trajectory appears smooth or feasible.
Then, the model returns a structured response consisting of a numerical score from 1 to 10, along with a concise explanation. Scores are interpreted based on task success and quality: low scores (1–4) indicate failure to accomplish the task, mid-range scores (6–8) reflect successful but imperfect trajectories, and high scores (9–10) are reserved for trajectories that are both accurate and high-quality. This scoring scheme allows for nuanced, human-like evaluation that integrates both semantic understanding and visual reasoning.
By leveraging this multimodal prompt framework, GPT-score enables a robust and interpretable evaluation process that aligns closely with human judgment, overcoming the limitations of purely geometric or distance-based metrics. The complete prompt is presented as follows:

\begin{tcolorbox}[
colback=gray!10!white,
colframe=cyan!60!black,
title={Complete Prompt for GPT Score Metrics},
]
You are an expert evaluator in robotic manipulation and visual reasoning. Your job is to assess the quality of predicted trajectories based on task instructions and visual inputs.\\

You are given:

- A task instruction describing an object manipulation task.

- An image showing a predicted trajectory. \\

**Note:**

- In the image, the red circle indicates the start point, and the blue diamond indicates the end point.

- The trajectory represents the predicted movement path of the manipulated object, not the robot or end-effector.

- You should **evaluate the predicted trajectory as a proposed motion for the object that is supposed to be moved**, based on the task instruction — **not based on the static positions of objects in the image**. The objects have not actually moved. \\

**Evaluation Criteria (listed in order of importance):**

1. **Task Alignment and Success (most important)**  
   - Does the trajectory clearly and accurately fulfill the task instruction?  
   - **The trajectory must start at the correct location and end at a target position that aligns with the task goal.**  
   - Large deviations in the starting or ending point (e.g., wrong object, wrong destination, or stopping short of the goal) should result in a low score, even if the rest of the trajectory is smooth.  
   - If the task is not accomplished (due to incorrect goal interpretation or spatial execution), the score should be low regardless of other qualities.

2. **Feasibility**  
   - Is the movement physically plausible, smooth, and continuous?  
   - Are there any unrealistic discontinuities, sharp turns, or impossible transitions?  
   - Even if the movement is feasible, it should not receive a high score if the task is not completed.

3. **Obstacle Avoidance / Safety**  
   - Does the trajectory reasonably avoid collisions with surrounding objects?  
   - Minor risks may be tolerated if the task is completed successfully, but major or clear collisions should reduce the score. \\

**Scoring Guideline:**

- If the task is **not accomplished**, or if the start or end point is significantly incorrect, the score should typically be **4 or below**.

- If the task is completed but the trajectory has issues (e.g., roughness, minor risk of collision), a score in the **6–8** range is appropriate.

- A **score of 9–10** should be given only when the trajectory clearly completes the task, with good start/end accuracy, smooth motion, and reasonable safety. \\

Based on these criteria, provide a single overall score from 1 (very poor) to 10 (excellent), reflecting how well the task is accomplished. \\

Respond strictly in the following format:

Score: <1-10>

Explanation: <brief justification> \\

The task instruction is:  
\{task\_instruction\} \\

Please give your response.


\end{tcolorbox}

\begin{table}[htbp]
\caption{Details of the training data for \alg}
\label{tab:fsd_dataset}
\centering
\footnotesize
\begin{tabularx}{\textwidth}{l X X l X}
\toprule
\textbf{Stage} & \textbf{Task} & \textbf{Datasets} & \textbf{Samples} & \textbf{Data Sources} \\
\midrule
\multirow{10}{*}{Stage 1}
& GeneralQA (Caption, VQA, OCR, RegionVQA, Conversation, Grounding, Text)
& ShareGPT4V, VQAv2, OCR-VQA, Visual7W, ST-VQA, RefCOCO/+/g, VG, AS-Core, AS-V2, TextVQA, Visual7W
& 838k
& LLaVA~\cite{liu2024llavanext}, ASMv2~\cite{wang2025all} \\
\cmidrule(lr){2-5}
& General Spatial Reasoning
& KITTI, 2D3DS, ObjectRef, RegionRef, VSR, CLEVR, CLEVR-Math, SUPER-CLEVR, RAVEN
& 295k
& SpatialBot~\cite{cai2024spatialbot}, RoboPoint~\cite{yuan2024robopoint}, SAT~\cite{ray2024sat}, LLaVA-OneVision~\cite{li2024llava} \\
\cmidrule(lr){2-5}
& Embodied Spatial Reasoning
& FSD Level 1~(145k), Level 2~(86k) and Level 3~(19k)
& 250k
& FSD \\
\midrule
\multirow{6}{*}{Stage 2}
& Spatial Affordance Generation \& Understanding
& FSD Level 4~(24k)
& 24k
& FSD \\
\cmidrule(lr){2-5}
& Visual Trace Generation \& Understanding
& FSD Level 5~(26k)
& 26k
& FSD \\
\bottomrule
\end{tabularx}
\end{table}

\section{Details of Benchmarks and Baselines}
\label{app:benchmarks and baselines}

For \textbf{general spatial reasoning tasks}, the answers are typically multiple-choice questions with clear options. However, some spatial reasoning models show reduced instruction-following ability after fine-tuning, preventing them from directly outputting the correct option. To address this, we use a \textbf{lenient matching rule}, considering an answer correct if it includes either the correct content or the corresponding option.

For \textbf{object/region reference tasks}, we carefully fine-tuned and used a tailored prompt for each model. Most models, such as GPT-4o and ASMV2~\cite{wang2025all}, cannot directly output specific points. Similar to the validation process for RoboPoint~\cite{yuan2024robopoint}, we also found that using in-context learning to specify point output formats resulted in worse performance compared to directly outputting bounding boxes. Therefore, for these models, we instructed them to output bounding boxes directly. From these bounding boxes, we either uniformly sampled nine points or took the midpoint. Then, we calculated the proportion of points within the specified region to determine the final average accuracy.

For the \textbf{VABench-VisualTrace task}, due to a lack of strong baselines, we developed an additional Transformer-based prediction model, trained from scratch using the same data, which we named \textbf{DINOv2 Predictor}. In the DINOv2 Predictor, the visual encoder uses a pre-trained DINOv2~\cite{oquab2023dinov2}, encoding images to output a $(196, 768)$ feature vector. The language encoder uses a pre-trained T5-Base~\cite{Raffel2020ExploringTL}, outputting $(32, 768)$. These are concatenated with learnable embeddings $(8, 768)$ and passed through a Transformer encoder base. The output of the learnable embeddings is then read and passed through a linear layer to predict eight points. During training, we keep the language encoder fully frozen and train the visual encoder along with the other remaining parameters.

\section{More Visualizations and Examples}
\label{app:visualizations}

\begin{figure}[t]
\centering
\includegraphics[width=1\linewidth]{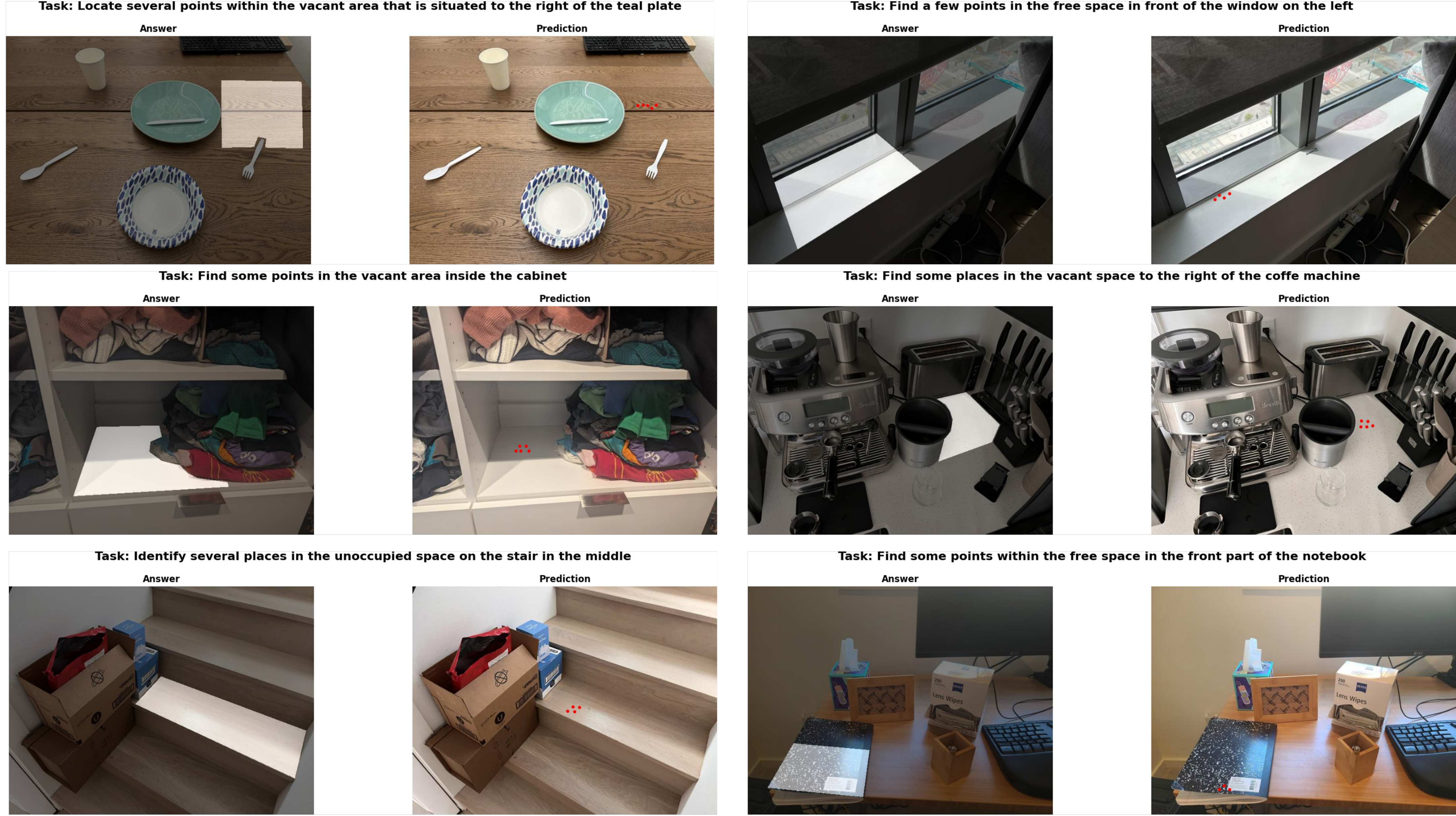}
\caption{\textbf{Visualization of visual aids generated by \alg in the Where2Place benchmark.}}
\label{fig:where2place}
\end{figure}

\begin{figure}[ht]
\centering
\includegraphics[width=0.8\linewidth]{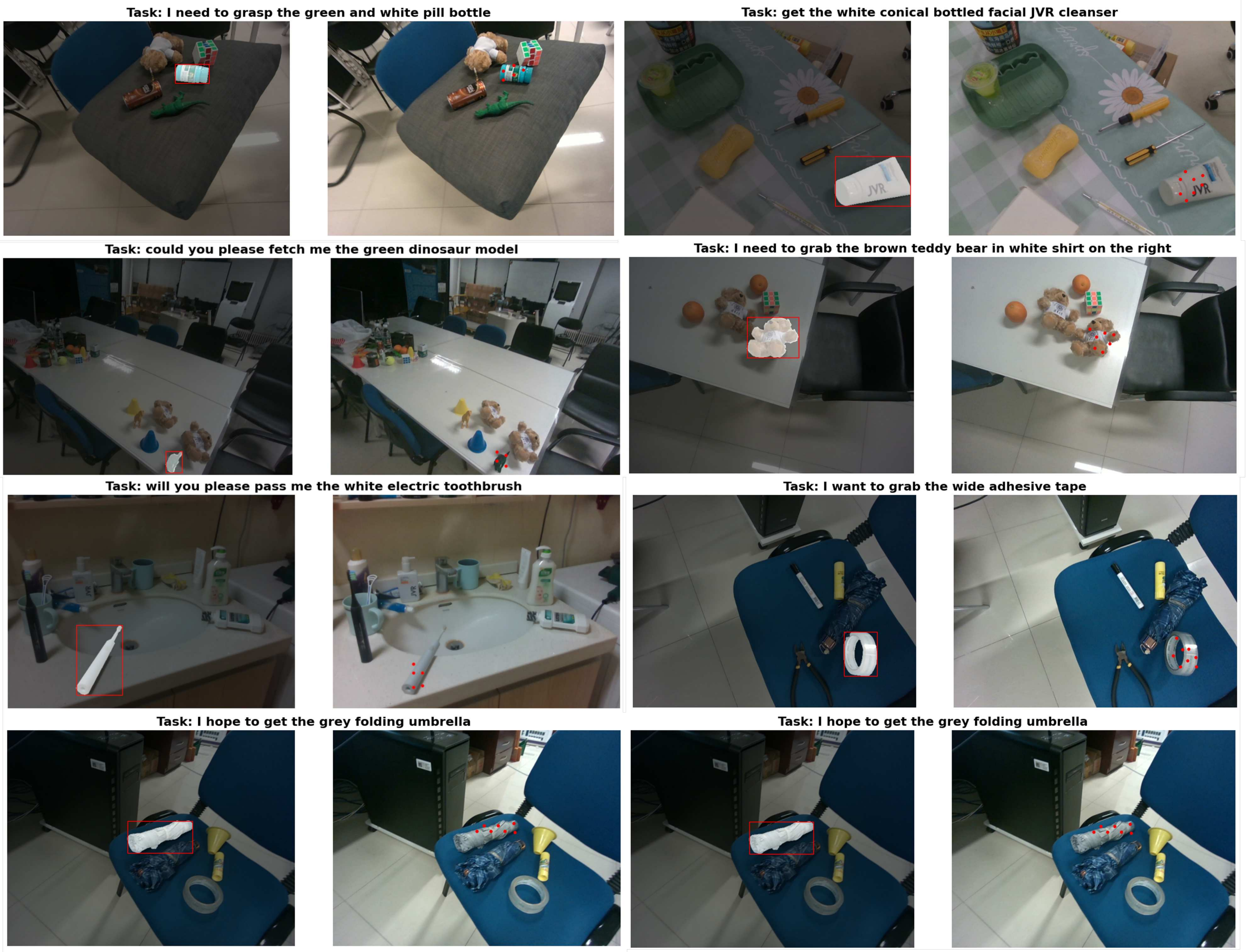}
\caption{\textbf{Visualization of visual aids generated by \alg in the RoboRefit benchmark.}}
\label{fig:roborefit}
\end{figure}

\begin{figure}[t]
\centering
\includegraphics[width=1\linewidth]{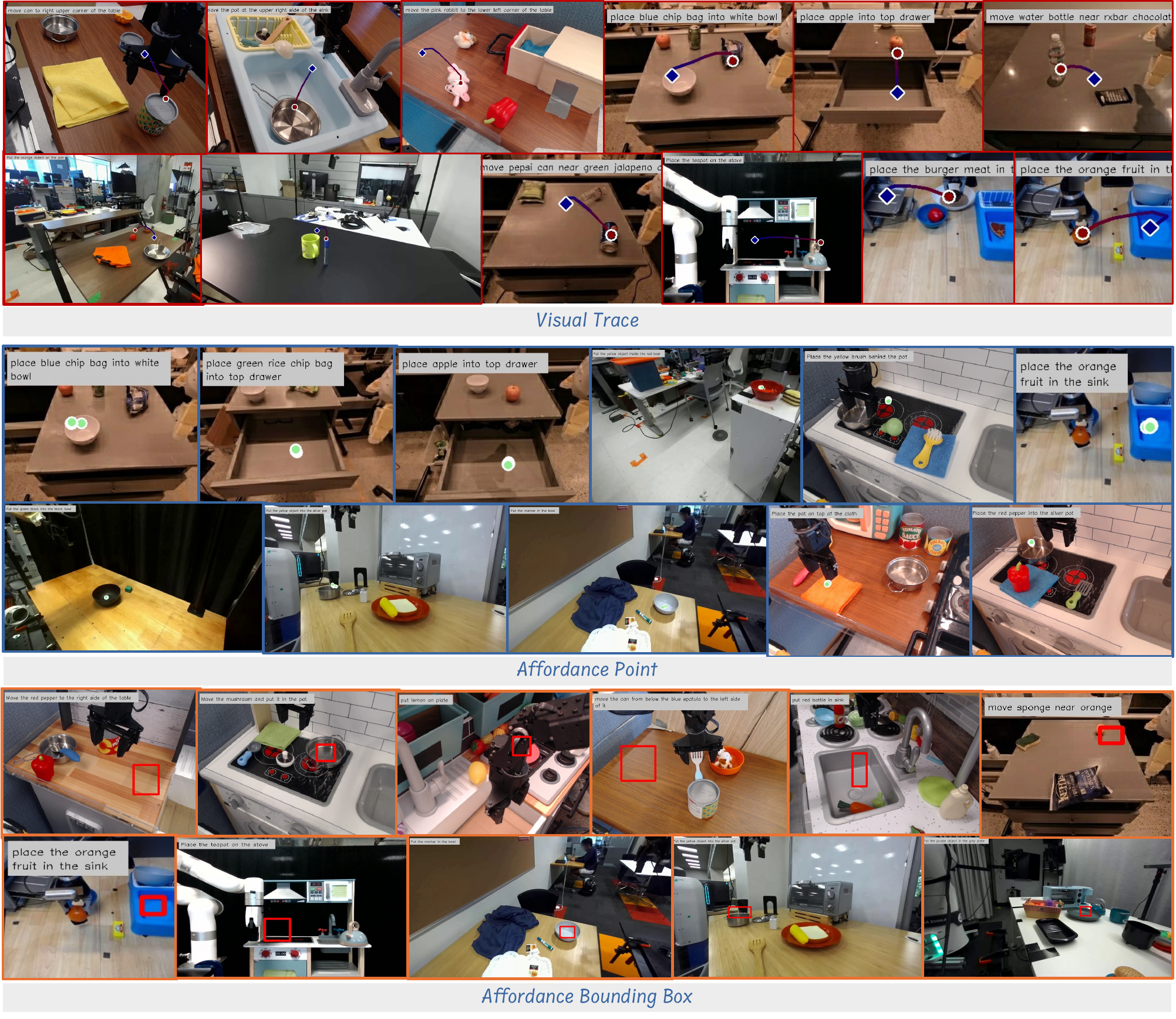}
\caption{\textbf{Visualization of visual aids generated by \alg in the VABench benchmark.} FSD can generate three types of visual aids based on task instructions for novel tasks and scenarios.
1st-2nd row: visual trace; 3rd-4th row: affordance points; 5th-6th rows: affordance bounding box.}
\label{fig:appendix_vis_aids}
\end{figure}

We present the prediction results of FSD on Where2place~\cite{yuan2024robopoint}, Roborefit~\cite{lu2023vl}, and VABench in \cref{fig:where2place}, \cref{fig:roborefit}, and \cref{fig:appendix_vis_aids}, respectively.

\section{Real World Experiment Results}
\label{app:real world experiements}

In our real-world desktop manipulation tasks, we used an xArm 6 robotic arm for evaluation. This setup included an Intel RealSense L515 LiDAR camera and a force-torque sensor on the xArm to enable compliance control, which improved interaction with the environment. A computer running Ubuntu 24.04 and equipped with an NVIDIA GTX 1660 was directly connected to the arm and camera to execute low-level control policies. Notably, a single RealSense L515 depth camera was sufficient for task completion, especially when performing visual trace execution. This approach eliminated the need for object segmentation and 3D mapping; instead, we directly mapped 2D visual trajectories to 3D for execution, with no strict requirements on depth information accuracy. Demonstrations are available in \cref{fig:montage} and on our \href{https://embodied-fsd.github.io/}{website}.

\section{Prompt for Using \alg Model}
\label{app:prompt for fsd}

\begin{tcolorbox}[title={Generate Spatial Affordance Points \& Bounding Box}]

You are currently a robot performing robotic manipulation tasks. Your task instruction: \{\texttt{Task Instruction}\} . Observe the image, use 2D points and bounding box to mark the target location where the manipulated object will be moved. In your answer, use <box>[[x1, y1, x2, y2]]</box> to present the bounding box of the target region, and use <point>[[x1, y1], [x2, y2], ...]</point> to mark the points of the free space.
\end{tcolorbox}

\begin{tcolorbox}[title={Generate Visual Trace}]

You are currently a robot performing robotic manipulation tasks. Your task instruction: \{\texttt{Task Instruction}\}. Observe the image, use 2D points to mark the manipulated object-centric waypoints to guide the robot to manipulate the object.Typically, the waypoints consists of an ordered sequence of eight 2D points. The format is <point>[[x1, y1], [x2, y2], ...]</point>.
\end{tcolorbox}

\section{Comparison of FSD and RoboBrain}
\label{app:fsd vs robobrain}

\begin{figure}[t]
\centering
\includegraphics[width=1\linewidth]{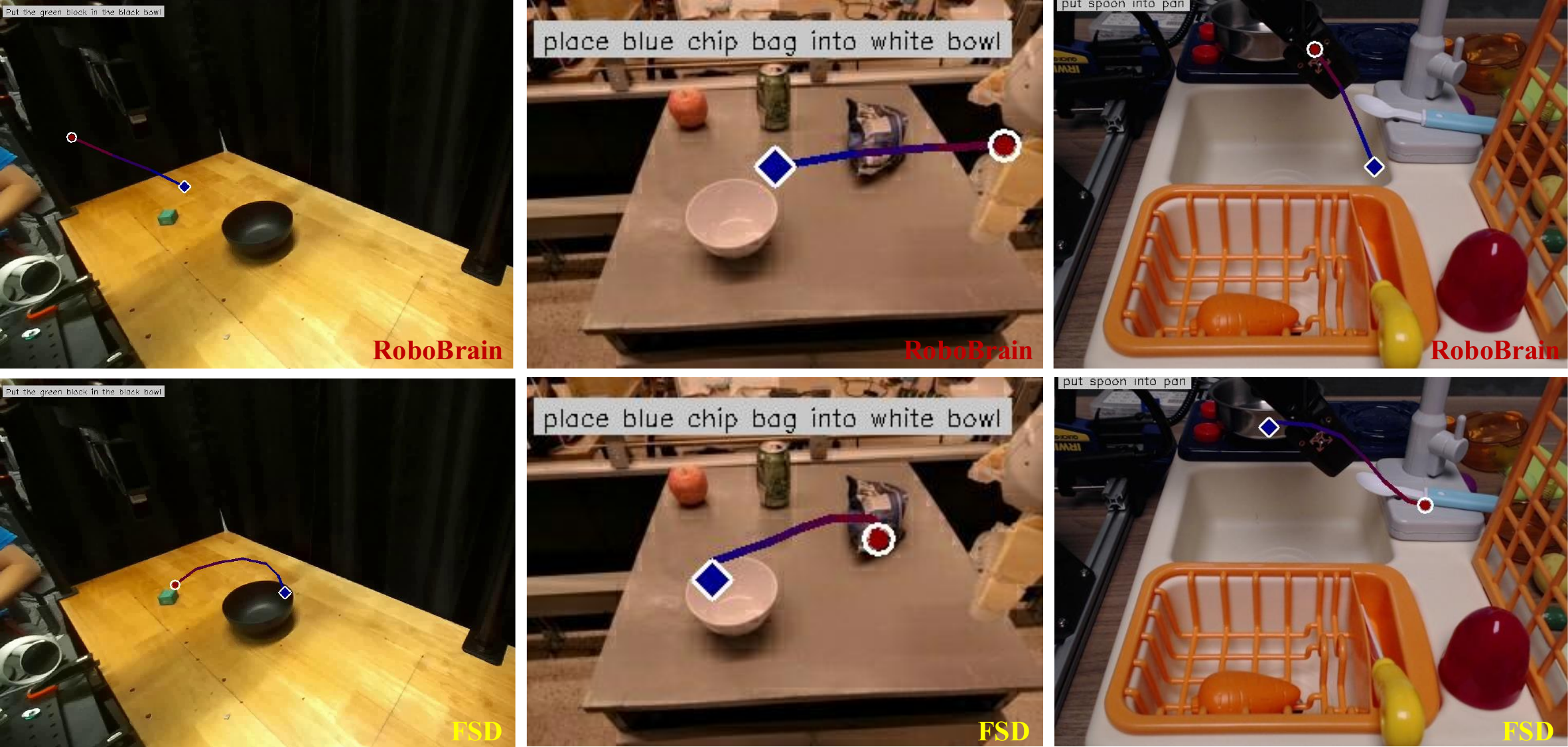}
\caption{\textbf{Comparison of generated visual traces between \alg and RoboBrain.}}
\label{fig:robobrain_fsd_compare}
\end{figure}

Both FSD and RoboBrain~\cite{ji2025robobrain} have the capability to generate visual trace. However, RoboBrain tends to produce agent-centric visual trace, whereas FSD generates object-centric visual trace. FSD adopts a task-centric design principle, allowing it to perform effectively even in more heterogeneous ontological scenarios, including those that completely lack robotic arms in the image, thus exhibiting stronger generalizability. Due to the differences in the methods of generating visual trace, we conducted several sets of visual trajectory visualizations for qualitative analysis, as shown in \cref{fig:robobrain_fsd_compare}. Under the same zero-shot setting, the visual trace generated by FSD has higher accuracy compared to those generated by RoboBrain, confirming that FSD’s reasoning-based pipeline possesses greater generalizability when facing unknown tasks.

\section{Future Works}
\label{app:future works}

We have made preliminary attempts to use visual aids as intermediate states in \alg, achieving promising results in object/target/region reference and actual manipulation task execution. Future work can focus on the following aspects to further enhance the applicability of this paradigm: 1. \textbf{Task Decomposition for Complex and Long-Horizon Instructions:} The current version of FSD primarily targets clear and explicit language instructions. When dealing with long-horizon tasks or ambiguous/complex instructions, the model needs to decompose them into atomic, executable sub-tasks. We believe that decomposing instructions into a sequence of visual aids to guide each sub-task execution is a promising avenue. 2.\textbf{Downstream Execution and Visual-Aid-Guided Control:}
Currently, FSD relies mainly on training-free motion planning methods for downstream execution. In extremely complex or dynamic scenarios, this may lead to a bottleneck in success rates. A potential improvement is to use the generated visual aids as explicit guidance for downstream VLA models, replacing language-conditioned training. Several preliminary studies~\cite{zheng2024tracevla, bharadhwaj2024track2act, li2025hamster} have shown that, for robotic manipulation tasks, affordance and visual trajectories provide more effective guidance than language prompts. 3. \textbf{Extending from 2D to 3D Visual Aids:} At present, FSD focuses on predicting 2D visual aids, similar to the representation used in Referring Expression Comprehension~(REC) and related tasks, which leverages the general visual understanding and reasoning capabilities of VLMs. However, as task and scene complexity increases, predicting 3D visual traces may prove to be a more effective solution, and we identify this as an important direction for future research.

\end{document}